\documentclass[10pt,twocolumn,letterpaper]{article}

\usepackage[]{cvpr} 
\usepackage{graphicx}
\usepackage{amsmath}
\usepackage{amssymb}
\usepackage{booktabs}
\usepackage[accsupp]{axessibility}
\usepackage[breaklinks,colorlinks]{hyperref}

\usepackage{algorithm}
\usepackage{algorithmic}
\usepackage{graphicx}
\usepackage{amsmath}
\usepackage{amssymb}
\usepackage{float}
\usepackage{color}
\usepackage{colortbl}
\usepackage{verbatim}
\usepackage{tabularx}
\usepackage{multirow}

\usepackage{color}
\usepackage{colortbl}
\usepackage{comment}

\usepackage[capitalize]{cleveref}
\crefname{section}{Sec.}{Secs.}
\Crefname{section}{Section}{Sections}
\Crefname{table}{Table}{Tables}
\crefname{table}{Tab.}{Tabs.}

\newcommand{\figref}[1]{Fig. \ref{#1}}
\newcommand{\tabref}[1]{Table \ref{#1}}

\makeatletter
\def\hlinewd#1{%
\noalign{\ifnum0=`}\fi\hrule \@height #1 \futurelet
\reserved@a\@xhline}

\begin{document}

\title{InstaFormer: Instance-Aware Image-to-Image Translation with Transformer}

\author{Soohyun Kim \quad Jongbeom Baek \quad Jihye Park \quad Gyeongnyeon Kim \quad Seungryong Kim\thanks{Corresponding author}\\
Korea University, Seoul, Korea\\
{\tt\small \{shkim1211,baem0911,ghp1112,kkn9975,seungryong\_kim\}@korea.ac.kr}
}

\maketitle

\begin{abstract}
We present a novel Transformer-based network architecture for instance-aware image-to-image translation, dubbed InstaFormer, to effectively integrate global- and instance-level information. By considering extracted content features from an image as tokens, our networks discover global consensus of content features by considering context information through a self-attention module in Transformers. By augmenting such tokens with an instance-level feature extracted from the content feature with respect to bounding box information, our framework is capable of learning an interaction between object instances and the global image, thus boosting the instance-awareness. We replace layer normalization (LayerNorm) in standard Transformers with adaptive instance normalization (AdaIN) to enable a multi-modal translation with style codes. In addition, to improve the instance-awareness and translation quality at object regions, we present an instance-level content contrastive loss defined between input and translated image. We conduct experiments to demonstrate the effectiveness of our InstaFormer over the latest methods and provide extensive ablation studies. 
\end{abstract}

\section{Introduction}

For a decade, image-to-image translation (I2I), aiming at translating an image in one domain (i.e., source) to another domain (i.e., target), has been popularly studied, to the point of being deployed in numerous applications, such as style transfer~\cite{gatys2016image,huang2017arbitrary}, super-resolution~\cite{dong2015image,kim2016accurate}, inpainting~\cite{pathak2016context,iizuka2017globally}, or colorization~\cite{zhang2016colorful, zhang2017real}. 

In particular, most recent works have focused on designing better disentangled representation to learn a multi-modal translation from unpaired training data~\cite{lee2018diverse,huang2018multimodal,park2020contrastive}. While they have demonstrated promising results, most of these methods only consider the translation on an \emph{whole} image, and do not account for the fact that an image often contain many object instances of various sizes, thus showing the limited performance at content-rich scene translation, e.g., driving scene, which is critical for some downstream tasks, such as domain adaptive object detection~\cite{bhattacharjee2020dunit}, that require well-translated object instances.

To address the aforementioned issues, some methods~\cite{shen2019towards,bhattacharjee2020dunit,jeong2021memory} seek to explicitly consider an object instance in an image within deep convolutional neural networks (CNNs). This trend was initiated by instance-aware I2I (INIT)~\cite{shen2019towards}, which treats the object instance and global image separately. Following this~\cite{shen2019towards}, 
some variants were proposed, e.g., jointly learning translation networks and object detection networks, called detection-based unsupervised I2I (DUNIT)~\cite{bhattacharjee2020dunit}, or using an external memory module, called memory-guided unsupervised I2I (MGUIT)~\cite{jeong2021memory}. While these methods improve an instance-awareness to some extent, they inherit limitation of CNN-based architectures~\cite{shen2019towards,bhattacharjee2020dunit,jeong2021memory}, e.g., local receptive fields or limited encoding of relationships or interactions between pixels or patches within an image which are critical in differentiating an object instance from an whole image and boosting its translation. 
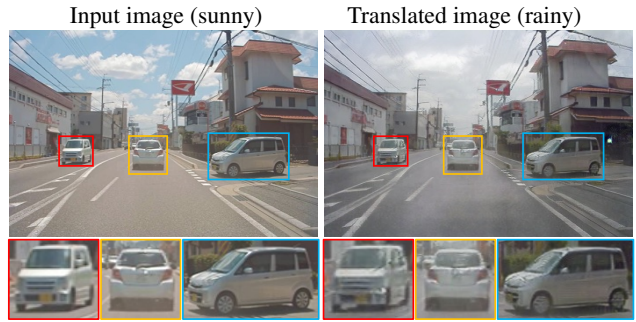
\begin{figure}[t]
\centering
	\begin{picture}(0,0)
	\put(-95,2){\small Input image (sunny)}
	\put(10,2){\small Translated image (rainy)}
	\end{picture}
\includegraphics[width=1\linewidth]{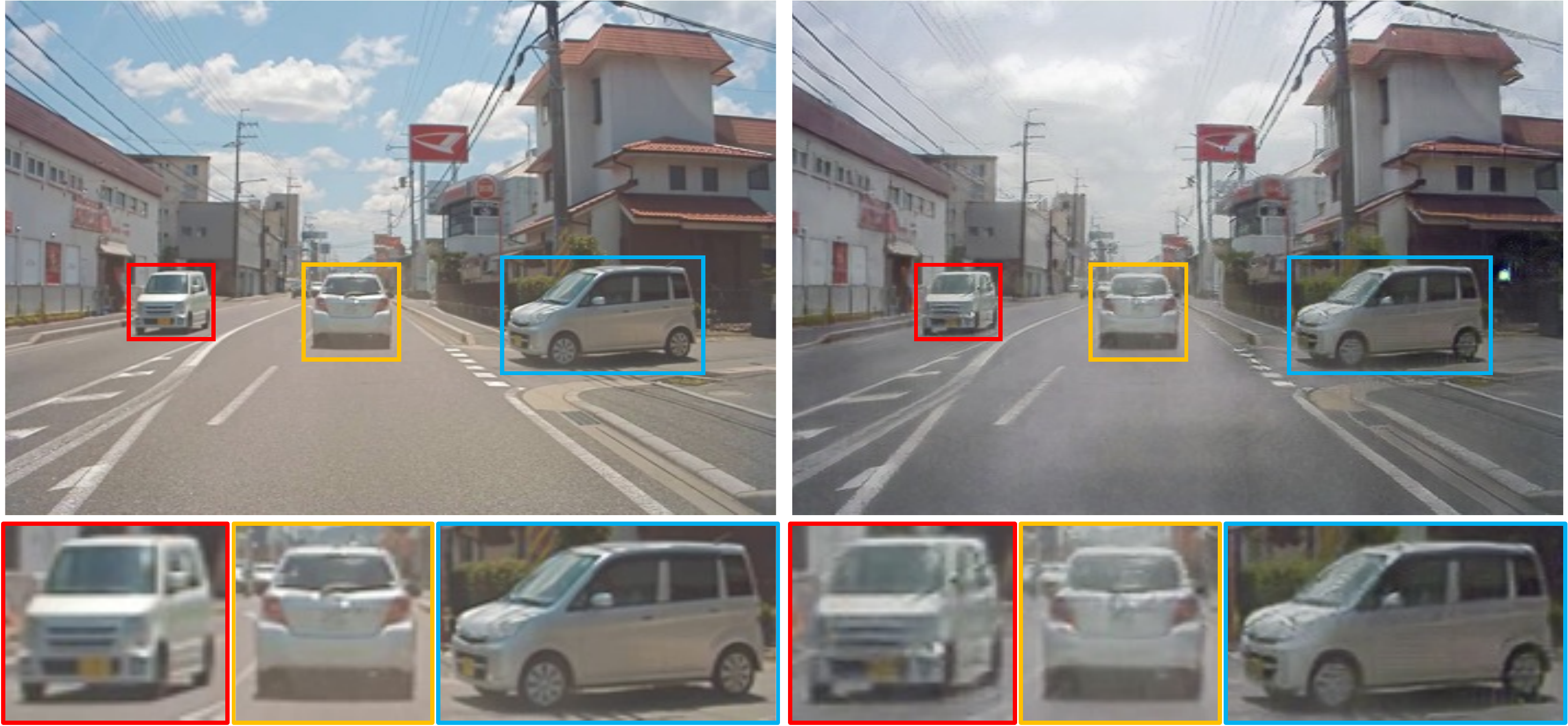}\hfill\\
\caption{
     \textbf{Results of InstaFormer for instance-aware image-to-image translation.} Our InstaFormer effectively considers global- and instance-level information with Transformers, which enables high quality instance-level translation.}
\label{f1}\vspace{-10pt}
\end{figure}

To tackle these limitations, for the first time, we present to utilize Transformer~\cite{vaswani2017attention} architecture within I2I networks that effectively integrates global- and instance-level information present in an image, dubbed InstaFormer. We follow common disentangled representation approaches~\cite{lee2018diverse,huang2018multimodal} to extract both content and style vectors. By considering extracted content features from an image as \textit{tokens}, our Transformer-based aggregator mixes them to discover global consensus by considering global context information through a self-attention module, thus boosting the instance-awareness during translation. In addition, by augmenting such tokens by an instance-level feature extracted from the global content feature with respect to bounding box information, our framework is able to learn an interaction between not only object instance and global image, but also different instances, followed by a position embedding technique to consider both global- and instance-level patches at once, which helps the networks to better focus on the object instance regions. We also replace layer normalization (LayerNorm)~\cite{ba2016layer} in Transformers with adaptive instance normalization (AdaIN)~\cite{huang2017arbitrary} to facilitate a multi-modal translation with extracted or random style vectors. Since aggregating raw content and style vectors directly with Transformers requires extremely large computation~\cite{dosovitskiy2020image,katharopoulos2020transformers}, we further propose to apply a convolutional patch embedding and deconvolutional module at the beginning and end of our Transformer-based aggregator. In addition, to improve the instance-awareness and quality of translation images at object regions, we present an instance-level content contrastive loss defined between input and translated images. 

In experiments, we demonstrate our framework on several benchmarks~\cite{geiger2012we,cordts2016cityscapes,shen2019towards} that contain content-rich scenes. Experimental results on various benchmarks prove the effectiveness of the proposed model over the latest methods for instance-aware I2I. We also provide an ablation study to validate and analyze components in our model.

\section{Related Work}
\paragraph{Image-to-Image Translation.}
While early efforts for I2I are based on supervised learning~\cite{isola2017image}, most recent state-of-the-arts focus on unpaired settings~\cite{zhang2020cross,zhou2021cocosnet,baek2021rethinking,zheng2021spatially,gabbay2021scaling,liu2021smoothing}. 
CycleGAN~\cite{zhu2017unpaired} attempts this by proposing a cycle-consistency loss which has been one of standard losses for unpaired I2I. Inspired by CycleGAN, numerous methods utilize cycle-consistency~\cite{kim2017learning,yi2017dualgan,choi2018stargan,hoffman2018cycada,lee2018diverse,huang2018multimodal}, and they can be largely divided into uni-modal models~\cite{liu2017unsupervised,yi2017dualgan,zhang2020cross} and multi-modal models~\cite{huang2018multimodal,lee2018diverse,choi2018stargan} methods. Specifically, MUNIT\cite{huang2018multimodal} assumes that an image representation can be disentangled into a domain-specific style and a domain-invariant content representation and uses these disentangled latent features with cycle-consistency to generate the translations. However, the content in translated image can be easily distorted, and cycle mapping requires multiple generators and discriminators. To address these, CUT~\cite{park2020contrastive}and F-LSeSim~\cite{zheng2021spatially} propose novel losses inspired by infoNCE~\cite{oord2018representation} to directly compute distance between input and translated images in an one-side framework without cycle-consistency. However, they still have shown limited performance to encode an object-awareness at the translated image.
\vspace{-10pt}

\paragraph{Instance-Aware Image-to-Image Translation.}
Some methods attempted to address the aforementioned issues~\cite{mo2018instagan,shen2019towards,bhattacharjee2020dunit,jeong2021memory}. INIT~\cite{shen2019towards} attempted to translate the whole image and object instances independently. DUNIT~\cite{bhattacharjee2020dunit} proposed to further train detection module and adopted instance-consistency loss for object-awareness.
MGUIT~\cite{jeong2021memory} utilizes bounding box to read and write class-wise memory module, and has access to class-aware features on memory at test-time. The aforementioned methods  inherit limitation of CNN-based architecture, e.g., local receptive fields or limited encoding of relationships or interactions within an image~\cite{zhu2017unpaired,lee2018diverse, huang2018multimodal}. 
\vspace{-10pt}

\paragraph{Vision Transformers and Image Generation}
Recently, Vision Transformers (ViT) have shown to attain highly competitive performance for a wide range of vision applications, such as image classification~\cite{dosovitskiy2020image,touvron2021training,wang2021pyramid,dai2021coatnet}, object detection~\cite{carion2020end,zhu2020deformable,dai2021up}, and semantic segmentation~\cite{zheng2021rethinking,xie2021segformer}. 
Inspired by ViT~\cite{dosovitskiy2020image}, some improvements are made to improve the computational complexity~\cite{katharopoulos2020transformers,wang2021pyramid,liu2021swin,wang2021pvtv2}. For example, Swin Transformer~\cite{liu2021swin} proposes relative position biases, and restricts self-attentioncomputation within shifted windows. MLP-Mixer~\cite{tolstikhin2021mlp} suggests to replace self-attention with an MLP, achieving memory efficiency and competitive performance~\cite{melas2021you,touvron2021resmlp,liu2021pay}. 
In this paper, we introduce ViT-based aggregator to further enhance to learn instance-awareness by aggregating information from local region as well as global image.

On the other hands, there exist several efforts to adapt Vision Transformers to image generation tasks~\cite{jiang2021transgan,zhao2021improved,chen2021pre,hudson2021generative,lee2021vitgan}. As seminal work, TransGAN~\cite{jiang2021transgan} first presents a GAN structure using pure Transformer, but has only validated on low-resolution images.~\cite{zhao2021improved} has achieved success on generating high-resolution images.~\cite{hudson2021generative} leverages Transformers to build the bipartite structure to allow long-range interactions. To our best knowledge, our work is the first attempt to adopt Transformers in instance-aware image translation.

\begin{figure*}
  \begin{subfigure}{0.78\linewidth}
  \centering
	{\includegraphics[width=1\linewidth]{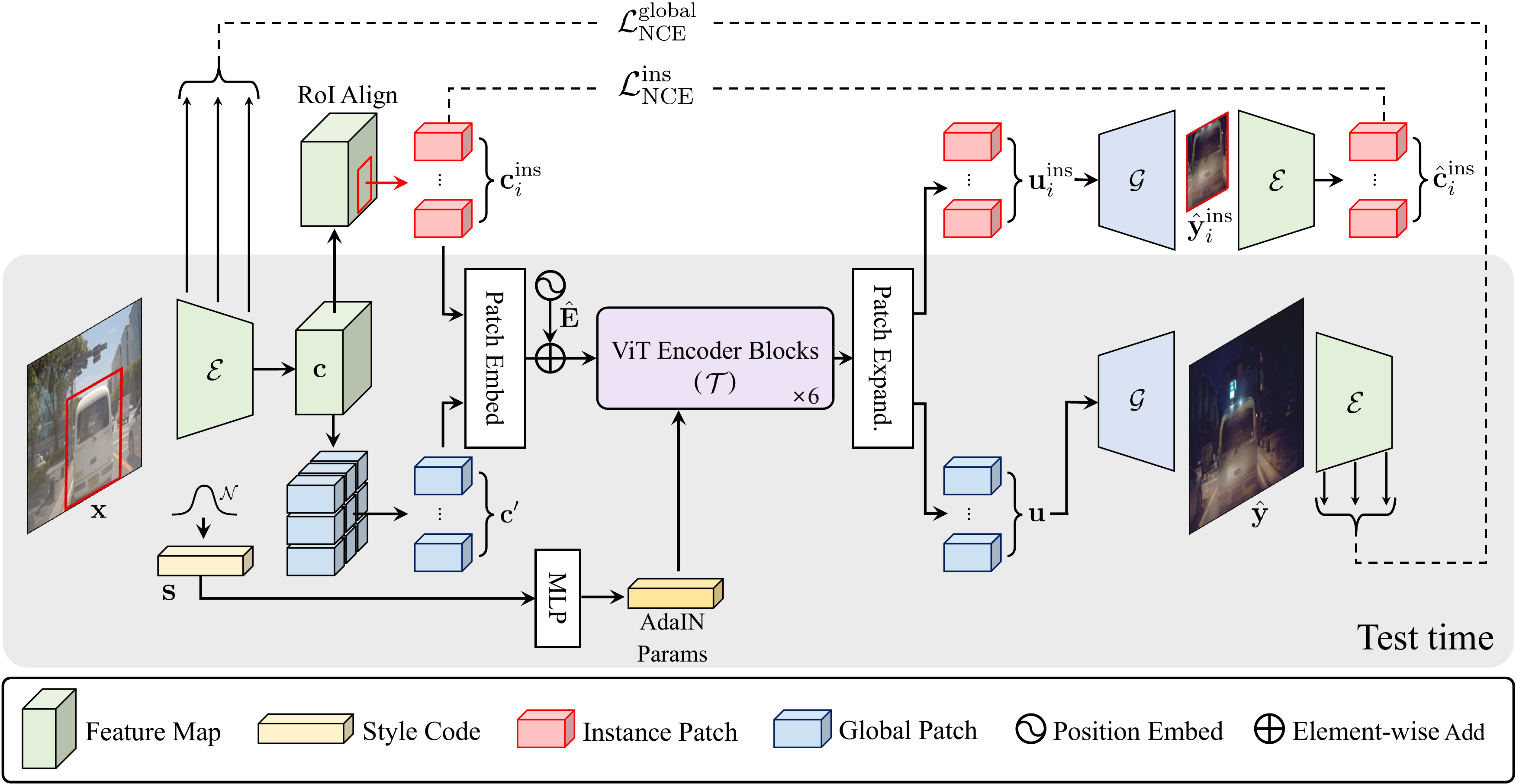}}\hfill
    \caption{Architecture}
  \end{subfigure}
  \hfill
  \begin{subfigure}{0.20\linewidth}
  \centering
	{\includegraphics[width=0.83\linewidth]{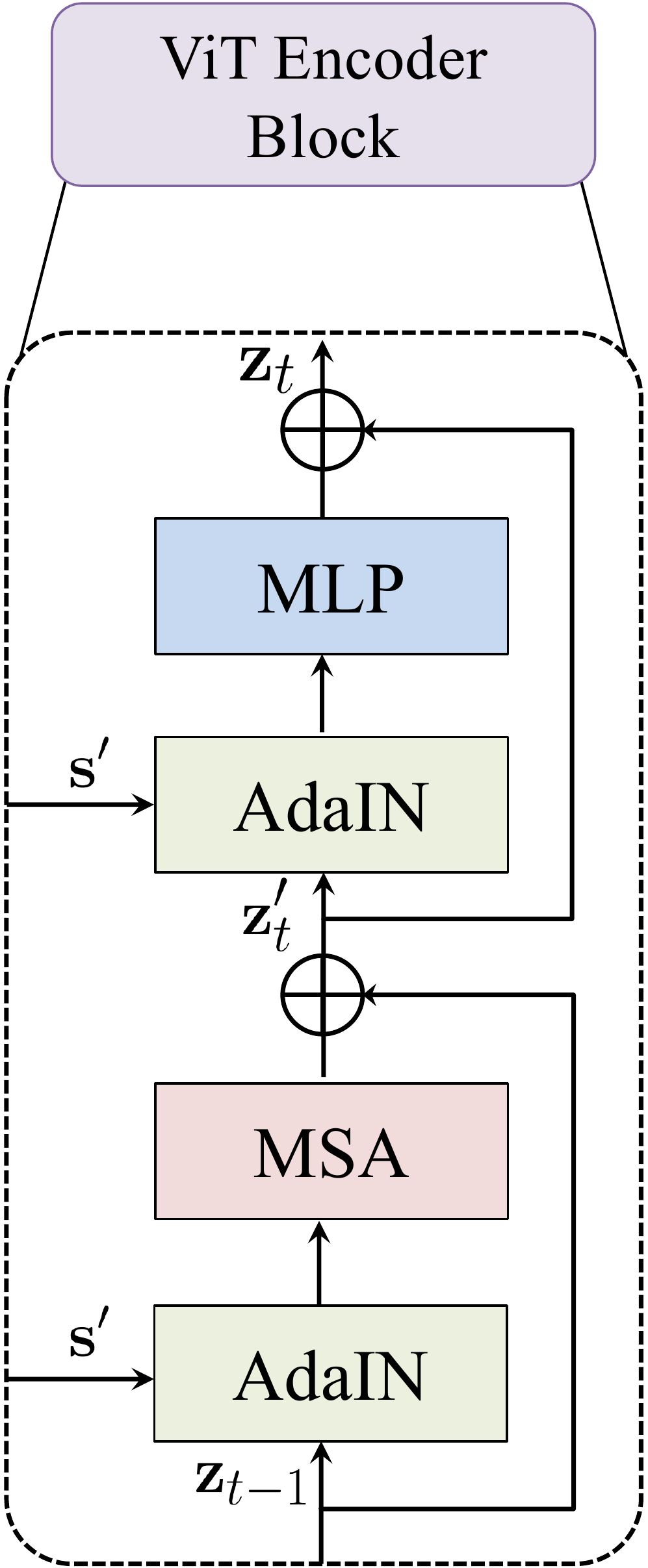}}\hfill\\
    \caption{ViT Encoder Block}
  \end{subfigure}
  \vspace{-5pt}
  \caption{\textbf{Network configuration:} (a) overall architecture for image-to-image translation, (b) ViT encoder block in details. Our networks consist of content encoder, Transformer encoder, and generator. The gray background represents the test phase, where we have no access on object instance bounding box (Best viewed in color).}
  \label{fig:architecture}

  \vspace{-10pt}
\end{figure*}
\section{Methodology}
\subsection{Overview}
\label{overview}
Our approach aims to learn a multi-modal mapping between two domains $\mathcal{X} \subset \mathbb{R}^ {H \times W \times 3}$ and $\mathcal{Y} \subset \mathbb{R}^ {H \times W \times 3}$ without paired training data, but with a dataset of unpaired instances $\mathrm{X} = \{\mathbf{x} \in \mathcal{X}\}$ and $\mathrm{Y} = \{\mathbf{y} \in \mathcal{Y}\}$. 
Especially, we wish to model such a mapping function to have an ability that jointly accounts for whole image and object instances. 
Unlike conventional I2I methods~\cite{lee2018diverse, huang2018multimodal, gonzalez2018image, shen2019towards, bhattacharjee2020dunit,jeong2021memory} that were formulated in a two-sided framework to exploit a cycle-consistency constraint, which often generates some distortions on the translated images and requires  auxiliary networks for inverse mapping~\cite{zhu2017unpaired}, 
we formulate our approach in an one-sided framework~\cite{park2020contrastive}. 

In specific, as illustrated in~\figref{fig:architecture}, our framework, dubbed InstaFormer, consists of content encoder $\mathcal{E}$ and generator $\mathcal{G}$, similar to~\cite{park2020contrastive,zheng2021spatially}, and additional encoder $\mathcal{T}$ with Transformers~\cite{vaswani2017attention} to improve the instance-awareness by considering global consensus between whole image and object instances.
To translate an image $\mathbf{x}$ in domain $\mathcal{X}$ to domain $\mathcal{Y}$, our framework first extracts a content feature map $\mathbf{c} = \mathcal{E}(\mathbf{x}) \in {\mathbb{R}}^{ h \times w \times l_{c}}$ from $\mathbf{x}$, with height $h$, width $w$, and $l_{c}$ channels, and randomly draws a style latent code $\mathbf{s} \in \mathbb{R}^{1 \times 1 \times {l_{s}}} $ from the prior distribution $q(\mathbf{s})\sim \mathcal{N}(0, \mathbf{I})$ to achieve a multi-modal translation. Instead of directly feeding $\mathbf{c}$ and $\mathbf{s}$ to the generator $\mathcal{G}$, as done in the literature~\cite{lee2018diverse,huang2018multimodal}, we aggregate information in the content $\mathbf{c}$ to discover global consensus between the global image and object instances in a manner that we first extract an object instance content vector $\mathbf{c}^\mathrm{ins}_i$ for $i$-th object bounding box with parameters $B_i = [x_i,y_i,h_i,w_i]$, where $(x_i,y_i)$ represent a center point, and $h_i$ and $w_i$ represent height and width of the box and $i \in 1,...,N$ where $N$ is the number of instance, and then mix $\{\mathbf{c},\{\mathbf{c}^\mathrm{ins}_i\}_i,\mathbf{s}\}$ through the proposed Transformer module $\mathcal{T}$ to extract global embedding $\mathbf{u}$ and instance embedding $\mathbf{u}^\mathrm{ins}_i$, which are independently used to generate global-level translated image $\hat{\mathbf {y}} = \mathcal{G}(\mathbf{u}) \in \mathbb{R}^{h \times w \times 3}$ and instance-level translated images $\hat{\mathbf{y}}^\mathrm{ins}_i = \mathcal{G}(\mathbf{u}^\mathrm{ins}_i) \in \mathbb{R}^{h_i \times w_i \times 3}$. In our framework, during training, we have access to the ground-truth object bounding boxes, while we do not access them at test-time.

To train our networks, we first use an adversarial loss defined between a translated image $\hat{\mathbf{y}}$ and a real image $\mathbf{y}$ from $\mathcal{Y}$ with discriminators, and a global content contrastive loss defined between $\mathbf{x}$ and $\hat{\mathbf{y}}$ to preserve the global content. To improve the disentanglement ability for content and style, following~\cite{huang2018multimodal}, we also use both image reconstruction loss and style reconstruction loss by leveraging an additional style encoder for $\mathcal{Y}$. To improve the instance-awareness and the quality of translation images at object instance regions, we newly present an instance-level content contrastive loss between $\mathbf{x}$ and $\hat{\mathbf{y}}$.

\begin{figure*}[t!]
  \begin{subfigure}{0.164\linewidth}{\includegraphics[width=1\linewidth]{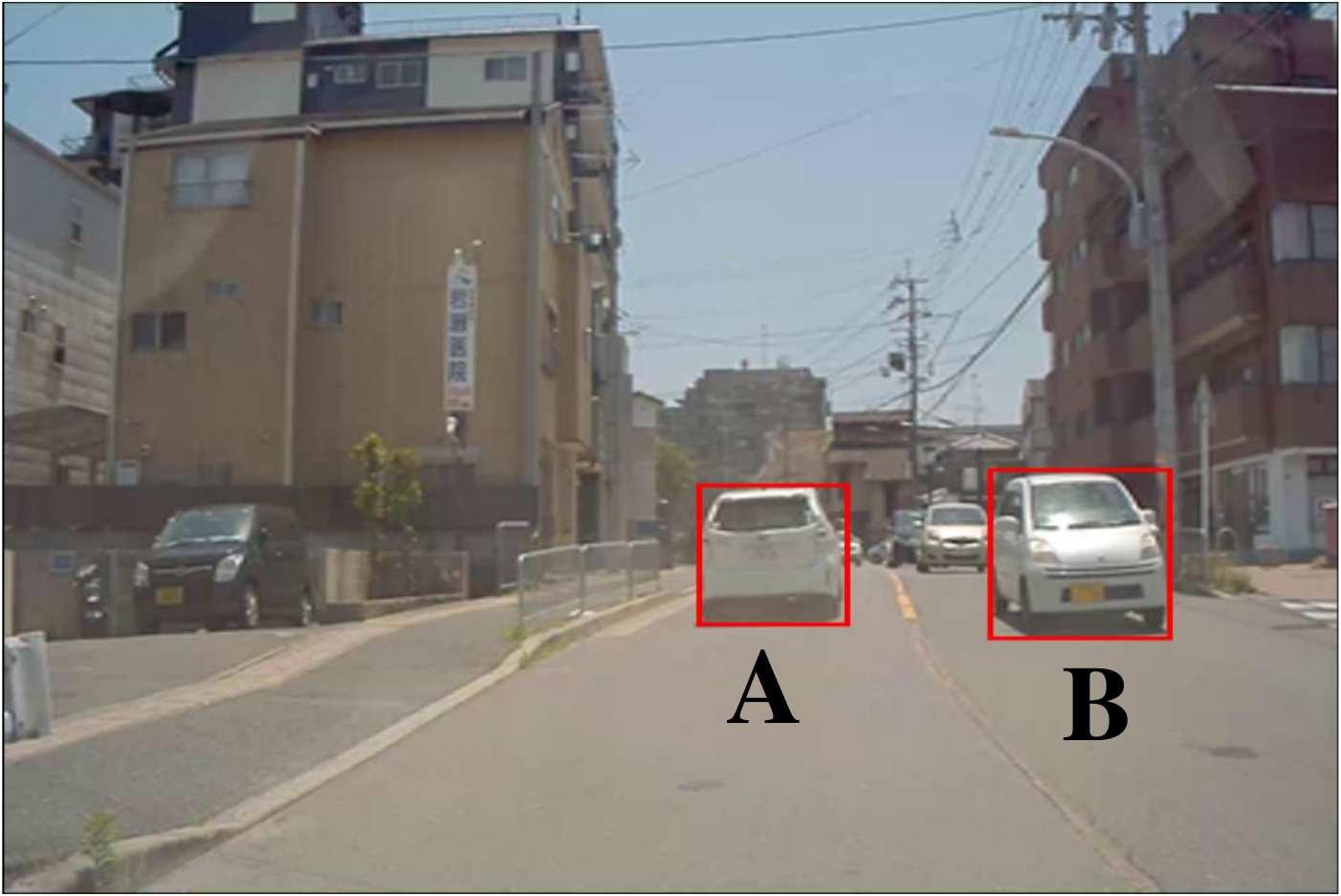}}
  \caption{content image}\end{subfigure}\hfill
  \begin{subfigure}{0.164\linewidth}{\includegraphics[width=1\linewidth]{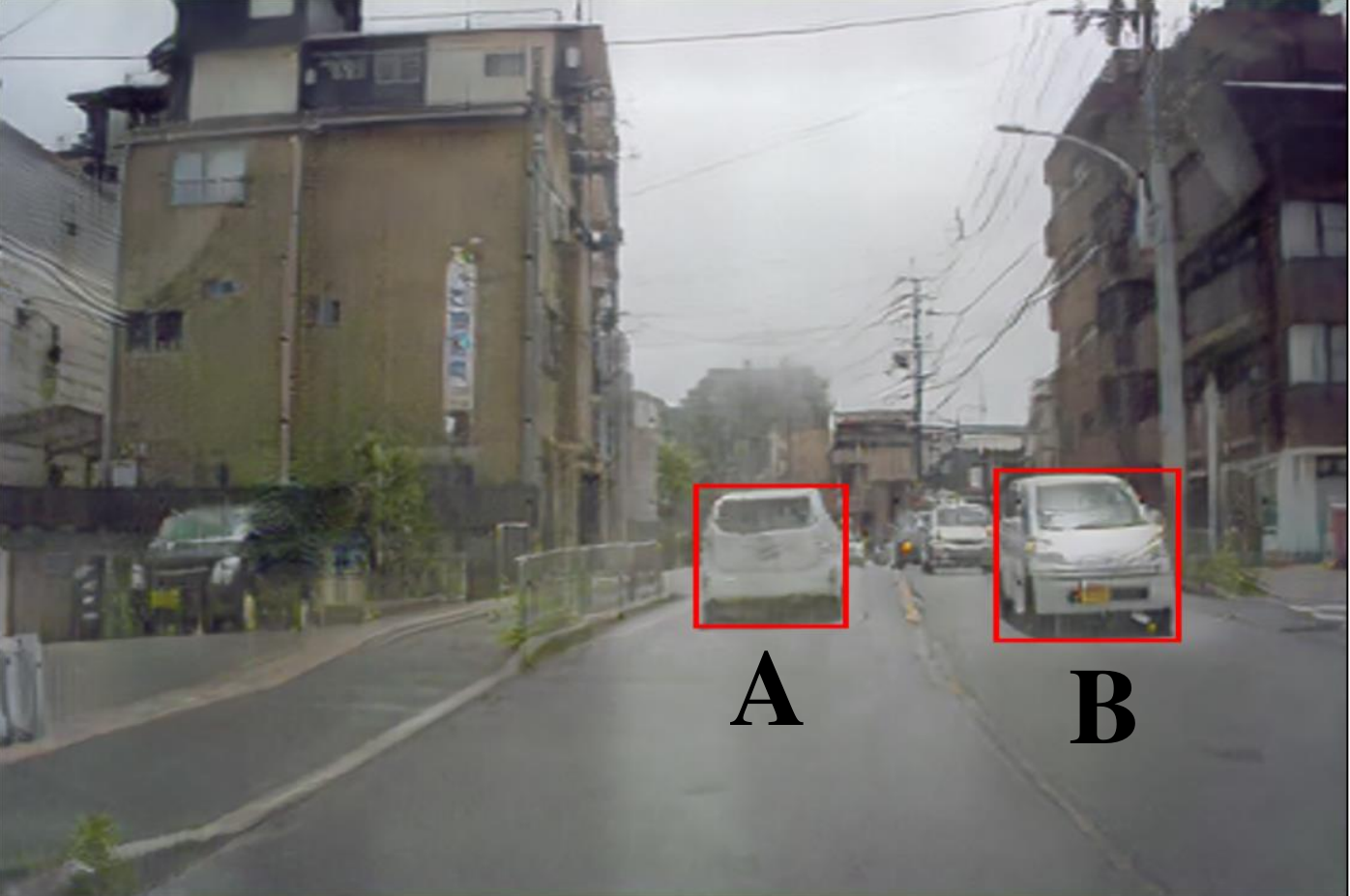}}
  \caption{translated image}\end{subfigure}\hfill
  \begin{subfigure}{0.164\linewidth}{\includegraphics[width=1\linewidth]{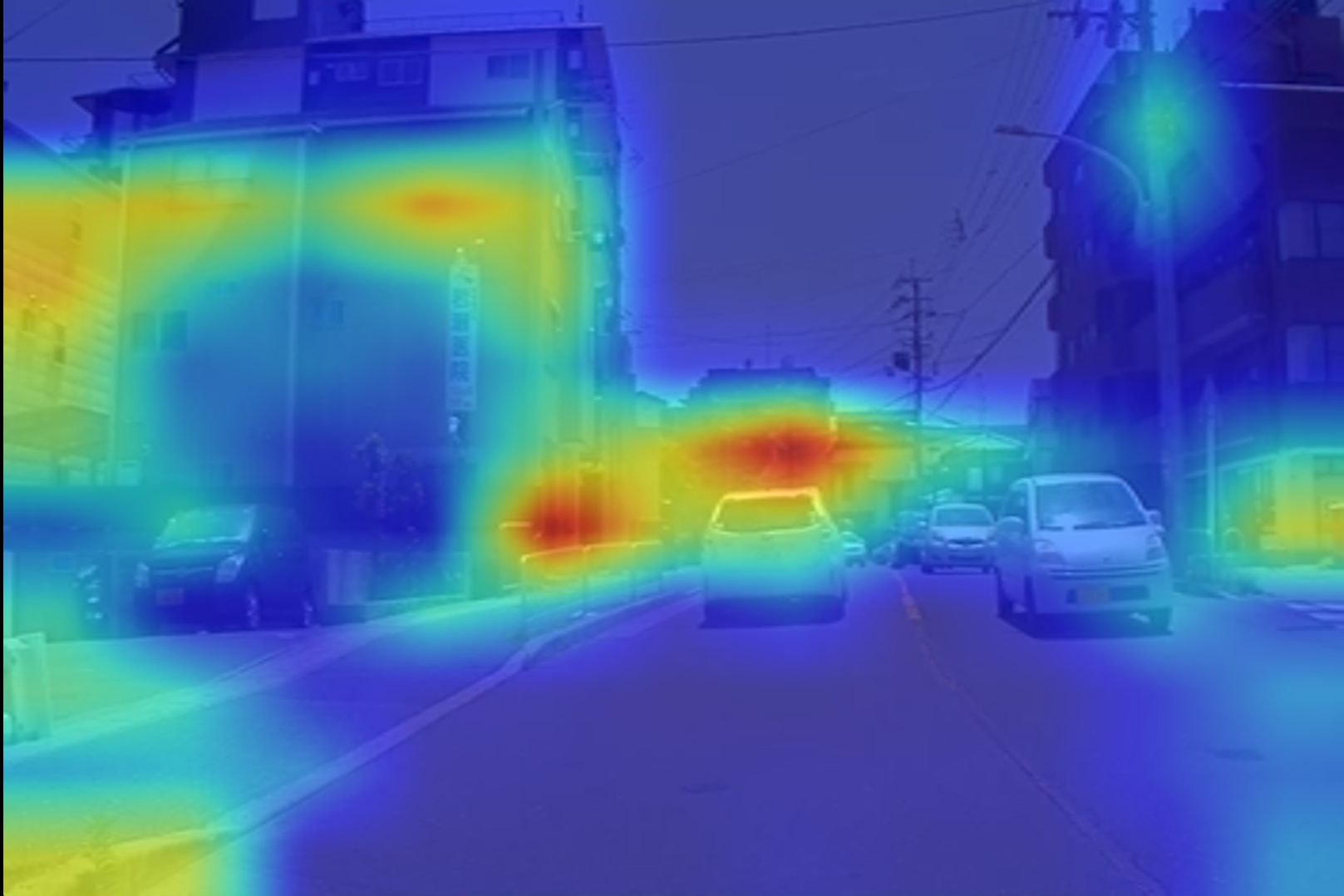}}
  \caption{w/o $\mathcal{L}_\mathrm{NCE}^\mathrm{ins}$ for A}\end{subfigure}\hfill
  \begin{subfigure}{0.164\linewidth}{\includegraphics[width=1\linewidth]{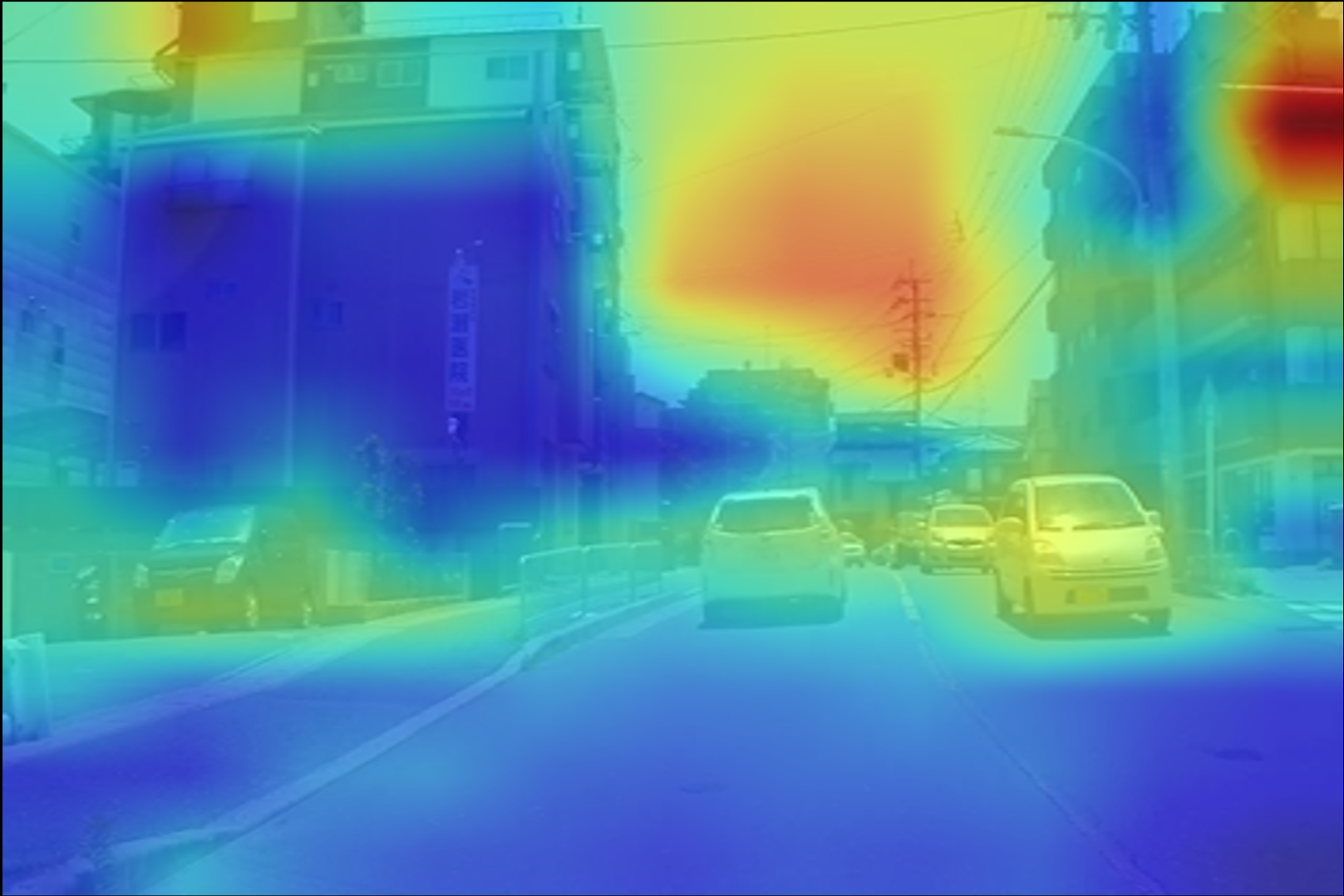}}
  \caption{w/o $\mathcal{L}_\mathrm{NCE}^\mathrm{ins}$ for B}\end{subfigure}\hfill
  \begin{subfigure}{0.164\linewidth}{\includegraphics[width=1\linewidth]{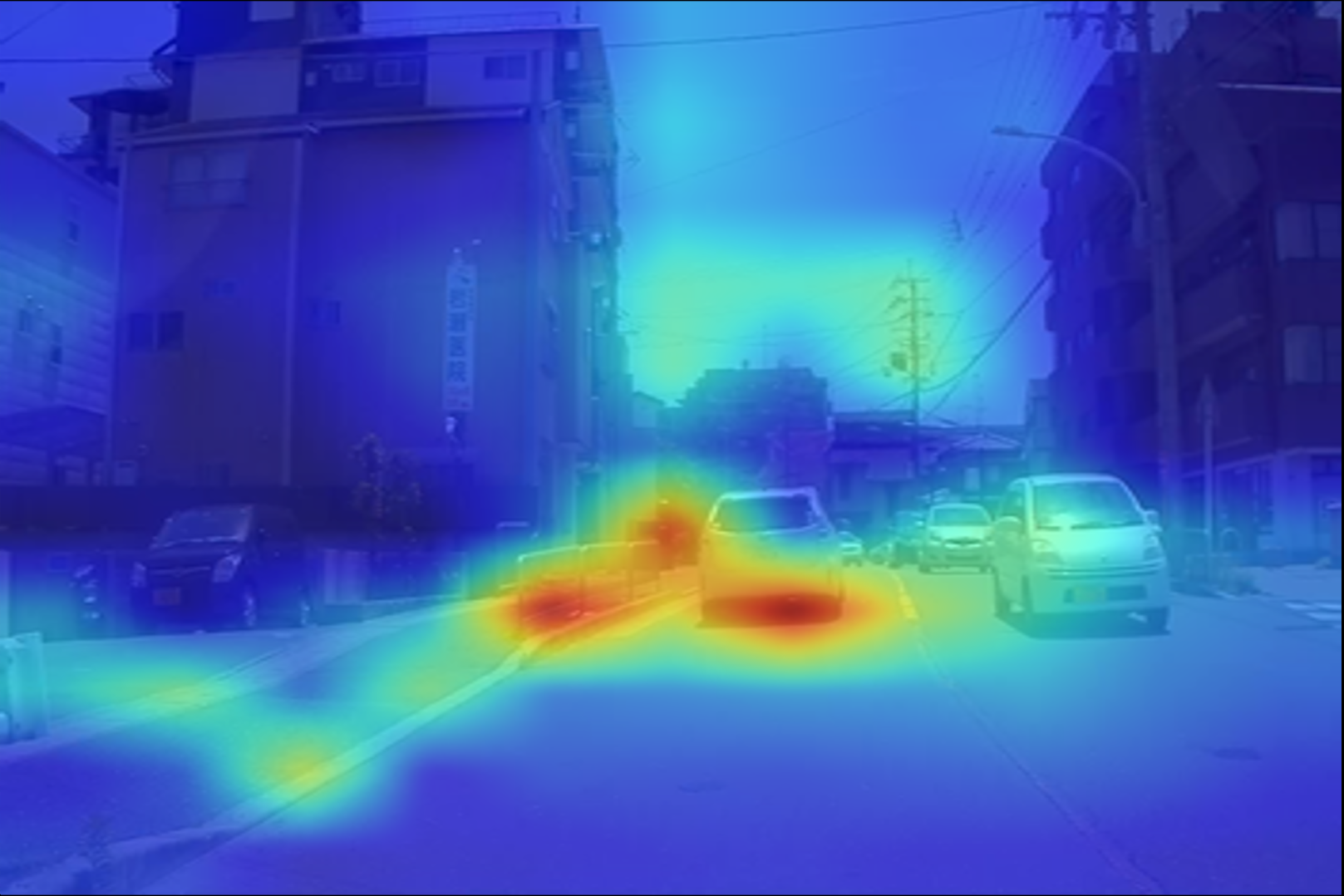}}
  \caption{w/ $\mathcal{L}_\mathrm{NCE}^\mathrm{ins}$ for A}\end{subfigure}\hfill
  \begin{subfigure}{0.164\linewidth}{\includegraphics[width=1\linewidth]{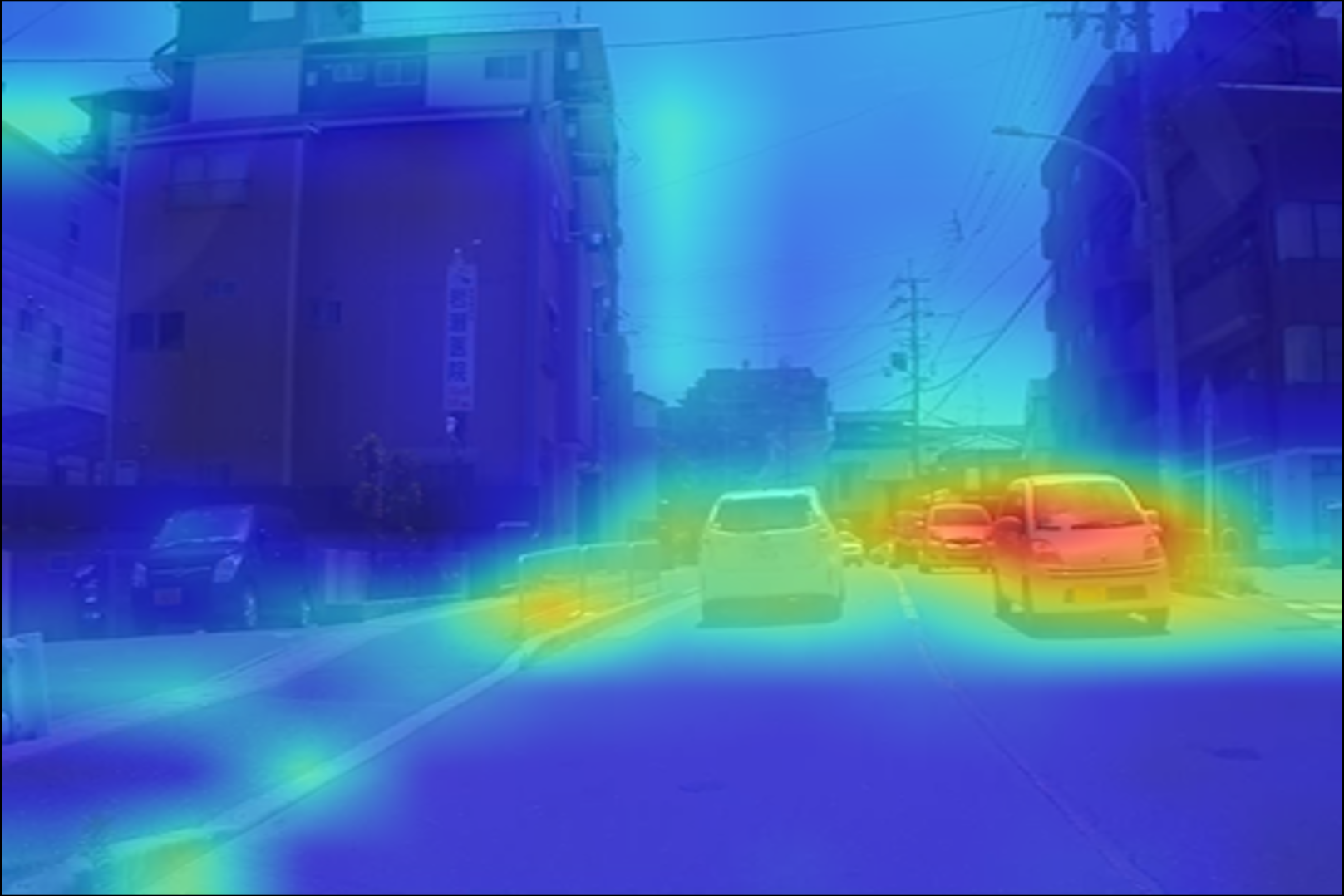}}
  \caption{w/ $\mathcal{L}_\mathrm{NCE}^\mathrm{ins}$ for B}\end{subfigure}\hfill  \\
\vspace{-5pt}
	\caption{\textbf{Visualization of learned self-attention.} For (a) content image containing instances A and B, our networks generate (b) translated image, considering attention maps (c,d) without $\mathcal{L}_\mathrm{NCE}^\mathrm{ins}$ and (e,f) with $\mathcal{L}_\mathrm{NCE}^\mathrm{ins}$ for instance A, B, respectively.}
	\vspace{-10pt}
	\label{fig:attention}
\end{figure*}

\subsection{Content and Style Mixing with Transformers}\label{section:vitmlp}
Most existing I2I methods~\cite{lee2018diverse,huang2018multimodal,shen2019towards, park2020contrastive,bhattacharjee2020dunit,jeong2021memory} attempted to aggregate a content feature map with deep CNNs with residual connections, which are often called residual blocks, often inserted between encoder and generator networks. They are thus limited in the sense that they inherit limitation of CNN-based architecture, e.g., local receptive fields or limited encoding of relationships or interactions between pixels and patches within an image~\cite{lee2018diverse,huang2018multimodal,shen2019towards}. In instance-aware I2I task, enlarging the receptive fields and encoding an interaction between objects and global image may be of prime importance. For instance, if an image contains a \textit{car} object on the \textit{road}, using the context information of not only global background, e.g., road, but also other instances, e.g., other cars or person, would definitely help to translate the image more focusing on the instance, but existing CNN-based methods~\cite{lee2018diverse,huang2018multimodal, park2020contrastive} would limitedly handle this.     

To overcome this, we present to utilize Transformer architecture~\cite{vaswani2017attention} to enlarge the receptive fields and encode the interaction between features for instance-aware I2I. To this end, extracted content vector $\mathbf{c} \in {\mathbb{R}}^{ h \times w \times l_{c}}$ from $\mathbf{x}$ can be flattened as a sequence $\mathbf{c}'=\text{Reshpae}(\mathbf{c})$ with the number of tokens $hw$ and channel $l_{c}$, which can be directly used as input for Transformers. However, this requires extremely high computational complexity due to the huge number of tokens $hw$, e.g., full HD translation. \vspace{-10pt}

\paragraph{Patch Embedding and Expanding.}
To address this issue, inspired by a patch embedding in ViT~\cite{dosovitskiy2020image}, we first apply sequential convolutional blocks to reduce the spatial resolutions. Instead of applying a single convolution in ViT~\cite{dosovitskiy2020image} for extracting non-overlapping patches, we use sequential overlapped convolutional blocks to improve the stability of training while reducing the number of parameters involved~\cite{xiao2021early}. We define this process as follows:
\begin{equation}
    \mathbf{p} = \mathrm{Conv}(\mathbf{c}) \in {\mathbb{R}} ^ {(h/k) \times (w/k) \times {l'_{c}}},
\end{equation}
where $k \times k$ is the stride size of convolutions, and $l'_{c}$ is a projected channel size. 
After feed-forwarding Transformer blocks such that $\mathbf{z} = \mathcal{T}(\mathbf{p})  \in {\mathbb{R}} ^ {(h/k) \times (w/k) \times {l'_{c}}}$, downsampled feature map $\mathbf{z}$ should be upsampled again with additional deconvolutional blocks, which are symmetric architectures to the convolutions, defined as follows:
\begin{equation}
    \mathbf{u} = \mathrm{DeConv}(\mathbf{z}) \in \mathbb{R}^\mathnormal{h \times w \times {l_{c}}}.
\end{equation}

In addition, for a multi-modal translation, we leverage a style code vector $\mathbf{s} \in \mathbb{R}^{1 \times 1 \times {l_{s}}}$, and thus this should be considered during mixing with Transformers~\cite{vaswani2017attention}. Conventional methods~\cite{huang2017arbitrary,lee2018diverse,huang2018multimodal} attempted to mix content and style vectors using either concatenation~\cite{lee2018diverse} or AdaIN~\cite{huang2017arbitrary}. In our framework, by slightly changing the normalization module in Transformers, we are capable of simultaneously mixing content and style vectors such that $\mathcal{T}(\mathbf{p},\mathbf{s})$.

Any forms of Transformers~\cite{dosovitskiy2020image,zhang2020feature,tolstikhin2021mlp,liu2021swin,wang2021pyramid} can be considered as a candidate in our framework, and in experiments, ViT-like~\cite{dosovitskiy2020image} architecture is considered for $\mathcal{T}$. In the following, we explain the details of Transformer modules.\vspace{-10pt}

\paragraph{Transformer Aggregator.}
In order to utilize Transformer to process content patch embeddings $\mathbf{p}$, our work is built upon the ViT encoder, which is composed of an multi-head self-attention (MSA) layer and a feed-forward MLP with GELU~\cite{vaswani2017attention}, where normalization layers are applied before both parts. Especially, for I2I, we adopt AdaIN instead of LayerNorm~\cite{ba2016layer} to control the style of the output with the affine parameters from style vector $\mathbf{s}$ and to enable multimodal outputs. In specific, content patch embedding $\mathbf{p}$ is first reshaped, and  position embedding is achieved such that
\begin{equation}
    \mathbf{z}_0 = \mathrm{Reshape}({\mathbf{p}})+ {\mathbf {E} \in {\mathbb{R}}^ {(h/k \cdot w/k) \times l_{c}'}},
\end{equation}
where $\mathbf{E}$ represents a position embedding~\cite{vaswani2017attention}, which will be discussed in the following. 

These embedded tokens $\mathbf{z}_0$ are further processed by the sequential Transformer encoder blocks as follows:
\begin{equation}
\begin{split}
    &\mathbf{z}'_{t} = \mathrm{MSA}\left( {\mathrm{AdaIN}\left({\mathbf{z}}_{t-1}, \mathbf{s}' \right)} \right) + {\mathbf{z}}_{t-1},\\
    &{\mathbf{z}}_t = \mathrm{MLP}\left( {\mathrm{AdaIN}\left(\mathbf{z}'_{t}, \mathbf{s}' \right)} \right) + \mathbf{z}'_{t},
\end{split}
\end{equation}
where $\mathbf{z}'_{t}$ and ${\mathbf{z}}_{t}$ denote the output of MSA and MLP modules for $t$-th block respectively and $t \in 1,...,T$, respectively, $\textbf{s}'$ indicates AdaIN parameters extracted from $S$. After $L$ Transformer modules, followed by reshaping to original resolution, we finally achieve the output of Transformer block $\mathcal{T}$ such that $\mathbf{z}_{T} = \mathcal{T}(\mathbf{p},\mathbf{s}')$. As exemplified in~\figref{fig:attention}, our learned self-attention well considers the interaction between object instances and global image.

\subsection{Instance-Aware Content and Style Mixing}
So far we discussed a method for content and style mixing with Transformers~\cite{dosovitskiy2020image}. This framework can improve the translation quality especially at instance regions to some extent, but the nature of irregular shape of object instances may hinder the performance boosting of our framework. In particular, global-level aggregation itself is limited to capture details of a tiny object and it is not always guaranteed that an object is located in a single regular patch. 
To overcome this, we present a novel technique to aggregate \emph{instance}-level content features and \emph{global}-level content features simultaneously, which enables the model to pay more attention to the relationships between global scenes and object instances.

In specific, given ground-truth bounding boxes with parameters $B_i$, we extract instance-level content feature maps through through ROI Align~\cite{he2017mask} module defined as follows:
\begin{equation}
    {\mathbf{c}}^\mathrm{ins}_{i} = \mathrm{RoIAlign}({\mathbf{c}}; {B}_{i}) \in {\mathbb{R}}^{k \times k \times {l_{c}}},
\end{equation}
where $k \times k$ is a fixed spatial resolution. This can be further processed with the convolutional blocks as proposed above such that
\begin{equation}
    {\mathbf{p}}^\mathrm{ins}_{i} = \mathrm{Conv}({\mathbf{c}}^\mathrm{ins}_{i}) \in {\mathbb{R}}^{1 \times 1 \times {l'_{c}}}.
\end{equation}

In our framework, by concatenating ${\mathbf{p}}$ and ${\mathbf{p}}^\mathrm{ins}_{i}$, we build a new input for Transformer $\hat{\mathbf{z}}_0$ such that 
\begin{equation}
    \hat{\mathbf{z}}_0 = \mathrm{Reshape}(\mathrm{Cat}({\mathbf{p}},\{{\bf{p}}^\mathrm{ins}_{i}\}_{i}))+ \hat{\mathbf{E}} \in {\mathbb{R}}^ {(h/k \cdot w/k+N) \times {l'_{c}}},
\end{equation}
where $\mathrm{Cat}(\cdot,\cdot)$ denotes a concatenation operator and $\hat{\mathbf{E}}$ is a corresponding positional embedding. Transformer blocks are then used to process $\hat{\mathbf{z}}_0$ similarly to above to achieve $\hat{\mathbf{z}}_{T}$, which is decomposed into ${\mathbf{z}}_{T}$ and ${\mathbf{z}}^{\mathrm{ins}}_{T,i}$.

\begin{figure}[t]
\centering
\includegraphics[width=0.75\linewidth]{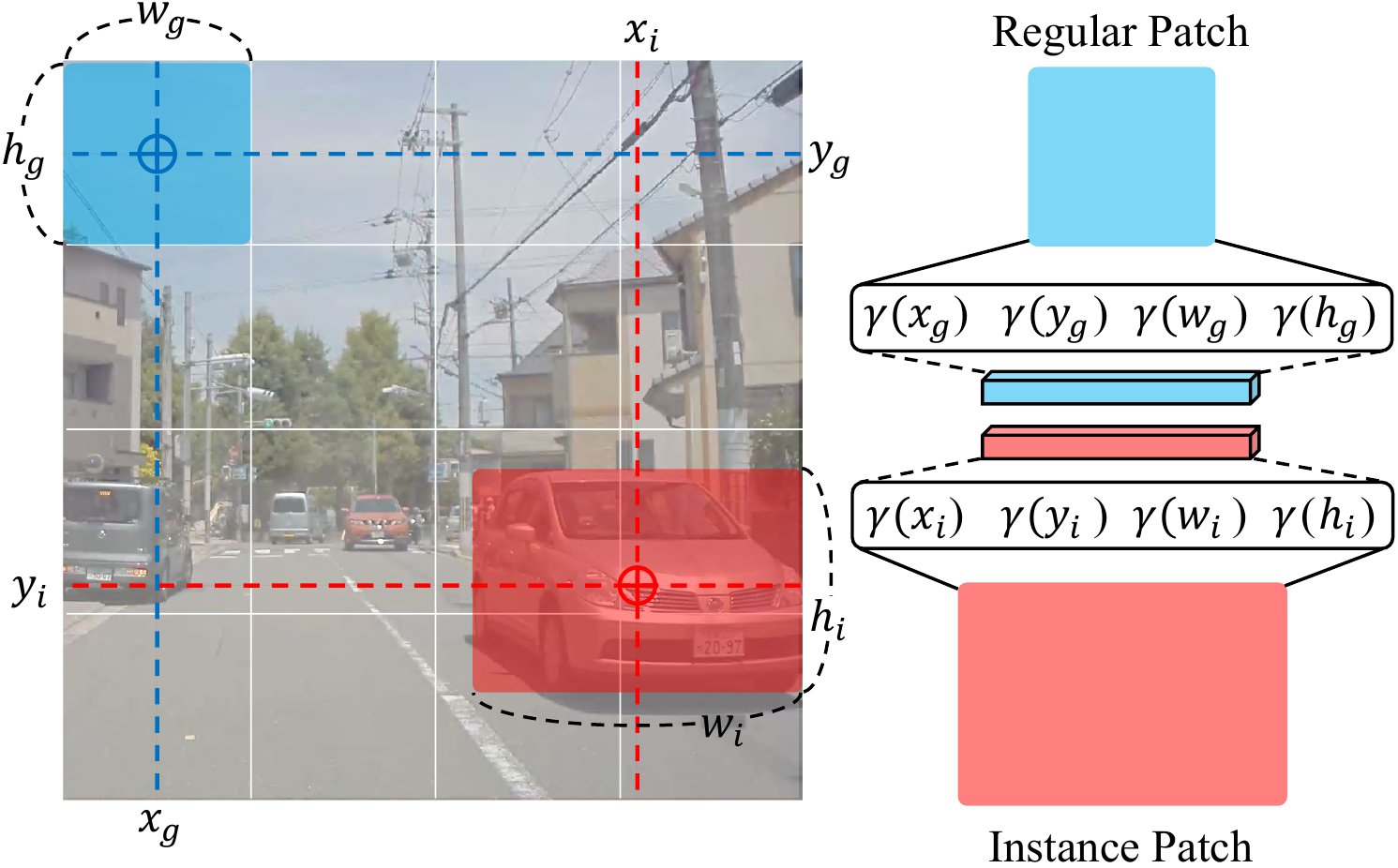}\hfill\\
\caption{\textbf{Illustration of building position embedding for regular patches and instance-level patches.}}
\vspace{-10pt}
\label{fig:pose}
\end{figure}

\subsection{Instance-Aware Position Embedding}
\label{sec:positional}
Since Transformer~\cite{vaswani2017attention} block itself does not contain positional information, we add positional embedding \textbf{E} as described above. To this end, our framework basically utilizes existing technique~\cite{dosovitskiy2020image}, but the main difference is that our proposed strategy enables simultaneously considering regularly-partitioned patches ${\mathbf{p}}$ and instance patches ${\mathbf{p}}^\mathrm{ins}_{i}$ in terms of their spatial relationships.

The deep networks are often biased towards learning lower frequency functions~\cite{rahaman2019spectral}, so we use high frequency functions to alleviate such bias.
We denote $\gamma(\cdot)$ as a sinusoidal mapping into $\mathbb{R}^{2K}$ such that $\gamma(\textit{a}) =(\mathrm{sin}(2^{0}\pi \textit{a}),\mathrm{cos}(2^{0}\pi \textit{a}),...,\mathrm{sin}(2^{K-1}\pi \textit{a}),\mathrm{cos}(2^{K-1}\pi \textit{a}))$ for a scalar $a$. 

In specific, as a global feature map is divided into regular girds, each regular patch can be represented to have center coordinates $(x_g, y_g)$ with patch width $w_g$ and height $h_g$ of regular size for $g$-th patch $\mathbf{p}(g)$. 
After embedding for each information through $\gamma(\cdot)$ and concatenating along the channel axis, it is further added to the patch embedded tokens. 
\begin{equation}
    \mathbf{E} = \mathrm{Cat}(\gamma(x_g),\gamma(y_g),\gamma(w_g),\gamma(h_g))
\end{equation}
Unlike regular patches, which have the same size of width and height for each, instance patches contain positional information of corresponding bounding boxes, which contain the centerpoint coordinates $(x_i, y_i)$ and width and height ($w_i$, $h_i$). Instance-wise $\mathbf{E}$ is denoted as:
\begin{equation}
\mathbf{E}^\mathrm{ins} = \mathrm{Cat}(\gamma(x_i),\gamma(y_i),\gamma(w_i),\gamma(h_i)).
\end{equation}
Then $\hat{\mathbf{E}}=\mathrm{Cat}(\mathbf{E},\mathbf{E}^\mathrm{ins})$.  \figref{fig:pose} illustrates the difference in how regular patch and instance patch are handled.

\begin{figure}[t]
\centering
\includegraphics[width=0.75\linewidth]{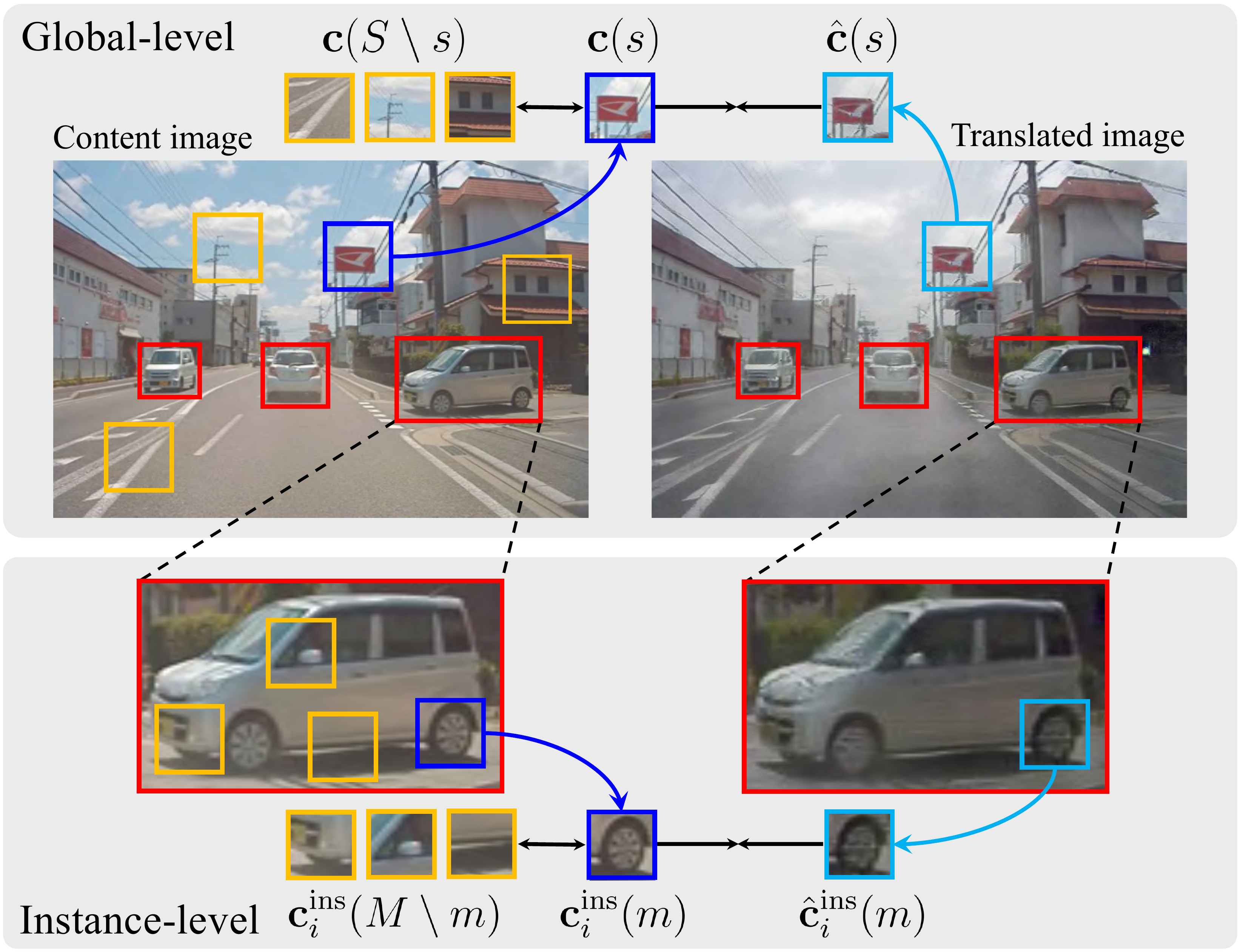}\hfill\\
\caption{\textbf{Illustration of global content loss and instance-level content loss.} Blue box indicates a positive sample, while yellow box means a negative sample (Best viewed in color).}
\vspace{-10pt}
\label{fig:loss}
\end{figure}

\subsection{Loss Functions}

\paragraph{Adversarial Loss.}
Adversarial loss aims to minimize the distribution discrepancy between two different features~\cite{goodfellow2014generative,mirza2014conditional}. We adopt this to learn the translated image $\hat{\mathbf{y}}$ to be similar to an image $\mathbf{y}$ from $\mathcal{Y}$ defined such that
\begin{equation}
\begin{split}
     \mathcal{L}_\mathrm{GAN} = &\mathbb{E}_{\mathbf{x}\sim\mathcal{X}}[\mathrm{log}(1-\mathcal{D}(\hat{\mathbf{y}}))]+ \mathbb{E}_{\mathbf{y} \sim\mathcal{Y}}[\mathrm{log}\, \mathcal{D}(\mathbf{y})],
\end{split}
\end{equation}
where $\mathcal{D}(\cdot)$ is the discriminator. \vspace{-10pt} 

\paragraph{Global Content Loss.}
To define the content loss between $\mathbf{x}$ and $\hat{\mathbf{y}}$, we exploit infoNCE loss~\cite{oord2018representation}, defined as
\begin{equation}
\begin{split}
    &\ell(\hat{\mathbf{v}}, \mathbf{v}^{+}, \mathbf{v}^{-}) =\\
    &\mathrm{-log}\left[\frac{\mathrm{exp}(\hat{\mathbf{v}}\cdot \mathbf{v}^{+}/\tau)}
    {\mathrm{exp}(\hat{\mathbf{v}}\cdot \mathbf{v}^{+}/\tau) + \sum_{\mathrm n=1}^{\mathrm N}\mathrm{exp}(\hat{\mathbf{v}}\cdot \mathbf{v}^{-}_{\mathrm n}/\tau)}\right],
\end{split}
\end{equation}
where $\tau$ is the temperature parameter, and $\mathbf{v^{+}}$ and $\mathbf{v^{-}}$ represent positive and negative for $\mathbf{\hat{v}}$. 

We set \emph{pseudo} positive samples between input image ${\mathbf{x}}$ and translated image $\hat{\mathbf{y}}$.
For the content feature from translated image $\hat{\mathbf{c}}(s)$ = $\mathcal{E(\hat{\mathbf {y}})}$, we set positive patches $\mathbf{c}(s)$, and negative patches $\mathbf{c}(S\setminus s)$ from $\mathbf{x}$, where $S\setminus s$ represents indexes except for $s$, following~\cite{park2020contrastive,zheng2021spatially}. Global content loss function is then defined as
\begin{equation}
\begin{split}
    \mathcal{L}_\mathrm{NCE}^{\mathrm{global}} = \mathbb{E}_{\mathbf{x}\sim\mathcal{X}}\sum_{l}\sum_{s}\ell(\hat{\mathbf{c}}_{l}(s),{\mathbf{c}}_{l}(s) , {\mathbf{c}}_{l}({S\setminus s})),
\end{split}
\end{equation}
where $\mathbf{c}_{l}$ is feature at $l$-th level, $s\in \{1,2,...,S_{l}\}$ and $S_{l}$ is the number of patches in each $l$-th layer.
\vspace{-10pt}

\begin{table*}[t!]
\centering
\scalebox{0.75}{
\begin{tabular}{l|cc|cc|cc|cc|cc|cc|cc|cc}
\hline
& \multicolumn{2}{c|}{CycleGAN~\cite{zhu2017unpaired}} & \multicolumn{2}{c|}{UNIT~\cite{liu2017unsupervised}} & \multicolumn{2}{c|}{MUNIT~\cite{huang2018multimodal}} & \multicolumn{2}{c|}{DRIT~\cite{lee2018diverse}} & \multicolumn{2}{c|}{INIT~\cite{shen2019towards}} & \multicolumn{2}{c|}{DUNIT~\cite{bhattacharjee2020dunit}} & \multicolumn{2}{c|}{MGUIT~\cite{jeong2021memory}} &
\multicolumn{2}{c}{InstaFormer}
\\ \cline{2-17}
& CIS & IS & CIS & IS & CIS & IS & CIS & IS & CIS & IS & CIS & IS & CIS & IS & CIS & IS\\ \hline\hline
sunny$\rightarrow$night	&	0.014	&	1.026	&	0.082	&	1.030	&	1.159	&	1.278	&	1.058	&	1.224	&	1.060	&	1.118	&	1.166	&	1.259	&	\underline{1.176}	&	\underline{1.271}	 & \textbf{1.200} & \textbf{1.404}\\
night$\rightarrow$sunny	&	0.012	&	1.023	&	0.027	&	1.024	&	1.036	&	1.051	&	1.024	&	1.099	&	1.045	&	1.080	&	\underline{1.083}	&	1.108	&	\textbf{1.115}	&	\textbf{1.130}	 & \textbf{1.115}  & \underline{1.127} \\\hline
sunny$\rightarrow$rainy	&	0.011	&	1.073	&	0.097	&	1.075	&	1.012	&	1.146	&	1.007	&	1.207	&	1.036	&	1.152	&	1.029	&	\underline{1.225}	&	\underline{1.092}	&	1.213	 & \textbf{1.158} & \textbf{1.394}\\
sunny$\rightarrow$cloudy	&	0.014	&	1.097	&	0.081	&	1.134	&	1.008	&	1.095	&	1.025	&	1.104	&	1.040	&	1.142	&	1.033	&	1.149	&	\underline{1.052}	&	\underline{1.218}	 & \textbf{1.130} & \textbf{1.257}\\
cloudy$\rightarrow$sunny	&	0.090	&	1.033	&	0.219	&	1.046	&	1.026	&	1.321	&	1.046	&	1.249	&	1.016	&	1.460	&	1.077	&	1.472	&	\underline{1.136}	&	\underline{1.489}	 & \textbf{1.141} & \textbf{1.585}\\\hline\hline
Average	&	0.025	&	1.057	&	0.087	&	1.055	&	1.032	&	1.166	&	1.031	&	1.164	&	1.043	&	1.179	&	1.079	&	1.223	&	\underline{1.112}	&	\underline{1.254}	 & \textbf{1.149} & \textbf{1.353} \\\hline
\end{tabular}}
\vspace{-5pt}
\caption{\textbf{Quantitative evaluation on INIT dataset~\cite{shen2019towards}.}
For evaluation, we perform bidirectional translation for each domain pair.
We measure CIS~\cite{huang2018multimodal} and IS~\cite{salimans2016improved}(higher is better).
Our results shows the best results in terms of CIS and IS.}
\vspace{-10pt}
\label{tab1}
\end{table*}

\paragraph{Instance-level Content Loss.} 
To improve the instance-awareness and the quality of translation images at object regions, we newly present an instance-level content contrastive loss. 

Our instance-level content loss is then defined such that
\begin{equation}
\begin{split}
\mathcal{L}_\mathrm{NCE}^\mathrm{ins}=\mathbb{E}_{\mathbf{x}\sim\mathcal{X}}\sum_{i}\sum_{m}\ell(\hat{\mathbf{c}}_{i}^{\mathrm{ins}}(m),{\mathbf{c}}_{i}^{\mathrm{ins}}(m) , {\mathbf{c}}_{i}^{\mathrm{ins}}({M\setminus m})),
\end{split}
\end{equation}
where $m\in\{1,2,...,M_{i}\}$ and $M_{i}$ is the number of patches at each instance. ~\figref{fig:loss} illustrates how our suggested content losses work, with the procedure to define positive and negative samples. 
\vspace{-10pt}

\paragraph{Image Reconstruction Loss.} 
We additionally make use of image reconstruction loss to help disentanglement between content and style. For regularization, we use a reconstruction loss to ensure that our $\mathcal{G}$ can reconstruct an image for domain $\mathcal{Y}$. To be specific, $\mathbf{y}$ is fed into $\mathcal{E}$ and style encoder $\mathcal{S}$ to obtain a content feature map $\mathbf{c^{\mathcal{Y}}}=\mathcal{E}(\mathbf{y})$ and a style code $\mathbf{s^{\mathcal{Y}}}=\mathcal{S}(\mathbf{y})$. 
We then compare the reconstructed image ${\mathcal{G}(\mathcal{T}(\mathbf{c^{\mathcal{Y}}},\mathbf{s^{\mathcal{Y}}}))}$ for domain $\mathcal{Y}$ with $\mathbf{y}$ as follows:
\begin{equation}
    \mathcal{L}_\mathrm{recon}^\mathrm{img} = \, \mathbb{E}_{\mathbf{y} \sim \mathcal{Y}}[\|{\mathcal{G}(\mathcal{T}(\mathbf{c^{\mathcal{Y}}},\mathbf{s^\mathcal{Y}})) -\,\mathbf{y}}\|_{1}].\vspace{-10pt}
\end{equation}

\paragraph{Style Reconstruction Loss.}
In order to better learn disentangled representation, we compute L1 loss between style code from the translated image and randomly generated style code in order to enable mapping generated style features to Gaussian distribution such that \begin{equation}
\begin{split}
    \mathcal{L}_\mathrm{recon}^\mathrm{style} = \mathbb{E}_{\mathbf{x}\sim\mathcal{X} ,\mathbf{y}\sim\mathcal{Y}}[\|{\mathcal{S}(\hat{\mathbf{y}})
    -\,{\mathbf{s}}}\|_{1}].\vspace{-10pt}
\end{split}
\end{equation}

\paragraph{Total Loss.}
The total loss function is as follows:
\begin{equation}
\begin{split}
\min\limits_{{\mathcal{E}},\mathcal{G},\mathcal{S}} \max\limits_\mathcal{D}\mathcal{L}(\mathcal{E},\mathcal{G},\mathcal{D}) = &\mathcal{L}_{\mathrm{GAN}}
+\lambda^{\mathrm{glob}}\mathcal{L}_\mathrm{NCE}^{\mathrm{global}} +\lambda^\mathrm{ins}\mathcal{L}_\mathrm{NCE}^{\mathrm {ins}}
\\&+\lambda^\mathrm{style}\mathcal{L}_\mathrm{recon}^\mathrm{style}
+\lambda^\mathrm{img} \mathcal{L}_\mathrm{recon}^\mathrm{img}, 
\end{split}
\end{equation}
where $\lambda^{\mathrm{glob}}$, $\lambda^\mathrm{ins}$, $\lambda^\mathrm{style}$, and $\lambda^\mathrm{img}$ are weights that control the importance of each loss.

\section{Experiments}
\subsection{Implementation Details} 
We first summarize implementation details in our framework. We conduct experiments using a single 24GB RTX 3090 GPU. Training datasets are resized to the size of 352×352. We employ an Adam optimizer for 200 epochs using a step decay learning rate scheduler. A batch size of 8, an initial learning rate of 2e-4. The number of NCE layers $L$ is 3. For the loss weights, we set as $\lambda^{\mathrm{glob}}=1$, $\lambda^\mathrm{ins}=1$, $\lambda^\mathrm{style}=10$, and $\lambda^\mathrm{img}=5$. 
As described above, we implement our framework with the most representative vision Transformer-based, i.e., ViT\cite{dosovitskiy2020image}, but we will show our framework works with MLP-Mixer\cite{tolstikhin2021mlp} in the following. We will make our code publicly available.

\subsection{Experimental Setup}
We conduct experiments on two standard datasets for instance-aware I2I, INIT dataset~\cite{shen2019towards} and KITTI-Cityscapes dataset~\cite{geiger2012we,cordts2016cityscapes}.
INIT dataset~\cite{shen2019towards} provides street scene images including 4 domain categories (sunny, night, rainy, cloudy) with object bounding box annotations for car, person, and traffic sign. We conduct translation experiments for sunny$\rightarrow$night, night$\rightarrow$sunny, sunny$\rightarrow$rainy, sunny$\rightarrow$cloudy, and cloudy$\rightarrow$sunny.
KITTI object detection benchmark~\cite{geiger2012we} and Cityscapes~\cite{cordts2016cityscapes}
dataset are used to evaluate domain adaptation for object detection on KITTI$\rightarrow$Cityscapes. KITTI contains 7,481 images for training and 7,518 images for testing with the bounding box annotations for 6 object classes.
Cityscapes dataset consists of 5,000 images with pixel-level annotations for 30 classes.

In this section, we compared our InstaFormer with recent state-of-the-art instance-aware I2I methods: INIT~\cite{shen2019towards}, DUNIT~\cite{bhattacharjee2020dunit}, MGUIT~\cite{jeong2021memory}, and several unsupervised image-to-image translation methods: CycleGAN~\cite{zhu2017unpaired}, UNIT~\cite{liu2017unsupervised}, CUT~\cite{park2020contrastive}, MUNIT~\cite{huang2018multimodal}, and DRIT~\cite{lee2018diverse}.

\begin{figure*}
  \centering
  \begin{subfigure}{0.14\linewidth}{\includegraphics[width=1\linewidth]{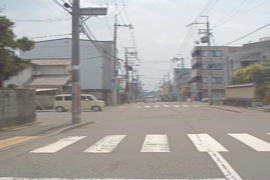}}
  \end{subfigure}\hfill
  \begin{subfigure}{0.14\linewidth}{\includegraphics[width=1\linewidth]{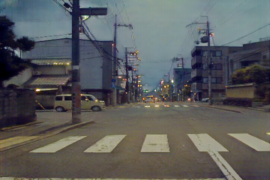}}
  \end{subfigure}\hfill
  \begin{subfigure}{0.14\linewidth}{\includegraphics[width=1\linewidth]{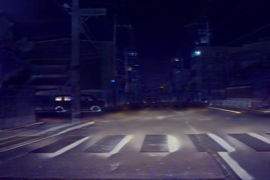}}
  \end{subfigure}\hfill
  \begin{subfigure}{0.14\linewidth}{\includegraphics[width=1\linewidth]{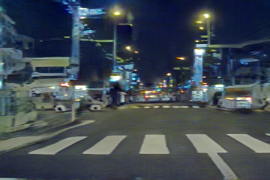}}
  \end{subfigure}\hfill
  \begin{subfigure}{0.14\linewidth}{\includegraphics[width=1\linewidth]{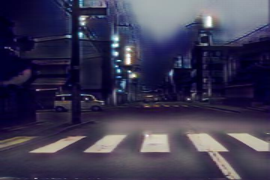}}
  \end{subfigure}\hfill
   \begin{subfigure}{0.14\linewidth}{\includegraphics[width=1\linewidth]{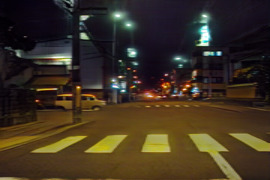}}
  \end{subfigure}\hfill
  \begin{subfigure}{0.14\linewidth}{\includegraphics[width=1\linewidth]{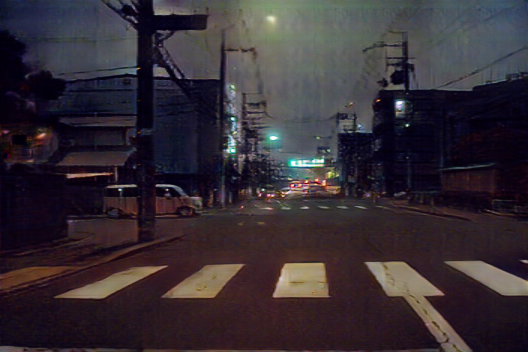}}
  \end{subfigure}\hfill\\
  
   \begin{subfigure}{0.14\linewidth}{\includegraphics[width=1\linewidth]{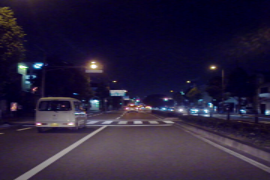}}
  \end{subfigure}\hfill
  \begin{subfigure}{0.14\linewidth}{\includegraphics[width=1\linewidth]{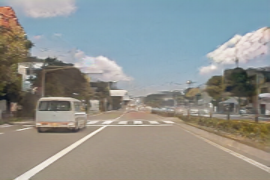}}
  \end{subfigure}\hfill
  \begin{subfigure}{0.14\linewidth}{\includegraphics[width=1\linewidth]{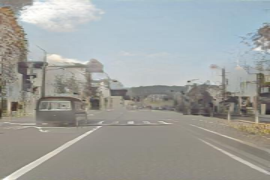}}
  \end{subfigure}\hfill
  \begin{subfigure}{0.14\linewidth}{\includegraphics[width=1\linewidth]{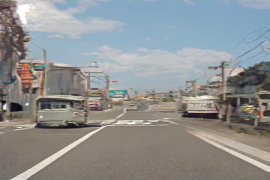}}
  \end{subfigure}\hfill
  \begin{subfigure}{0.14\linewidth}{\includegraphics[width=1\linewidth]{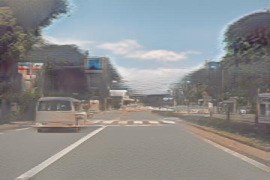}}
  \end{subfigure}\hfill
   \begin{subfigure}{0.14\linewidth}{\includegraphics[width=1\linewidth]{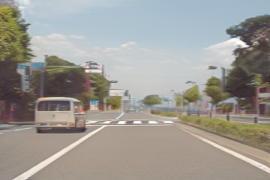}}
  \end{subfigure}\hfill
  \begin{subfigure}{0.14\linewidth}{\includegraphics[width=1\linewidth]{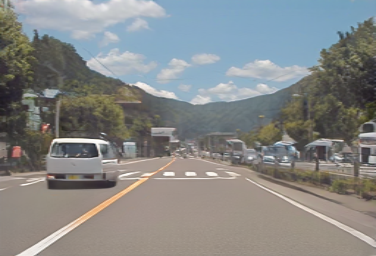}}
  \end{subfigure}\hfill\\

   \begin{subfigure}{0.14\linewidth}{\includegraphics[width=1\linewidth]{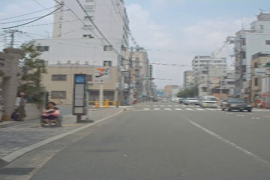}}
   \caption{Input}\end{subfigure}\hfill
  \begin{subfigure}{0.14\linewidth}{\includegraphics[width=1\linewidth]{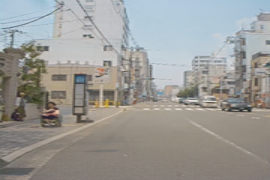}}
  \caption{CycleGAN~\cite{zhu2017unpaired}} \end{subfigure}\hfill 
  \begin{subfigure}{0.14\linewidth}{\includegraphics[width=1\linewidth]{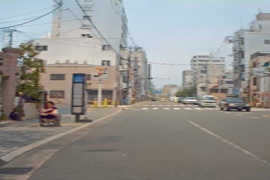}}
   \caption{UNIT~\cite{liu2017unsupervised}} \end{subfigure}\hfill
  \begin{subfigure}{0.14\linewidth}{\includegraphics[width=1\linewidth]{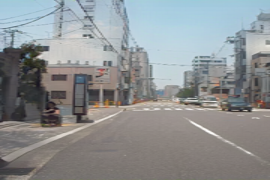}}
   \caption{MUNIT~\cite{huang2018multimodal}} \end{subfigure}\hfill
  \begin{subfigure}{0.14\linewidth}{\includegraphics[width=1\linewidth]{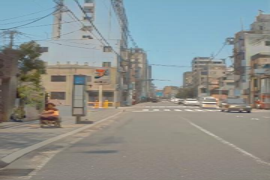}}
   \caption{DRIT~\cite{lee2018diverse}} \end{subfigure}\hfill
  \begin{subfigure}{0.14\linewidth}{\includegraphics[width=1\linewidth]{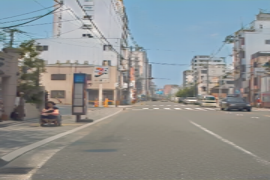}}
   \caption{MGUIT~\cite{jeong2021memory}} \end{subfigure}\hfill
  \begin{subfigure}{0.14\linewidth}{\includegraphics[width=1\linewidth]{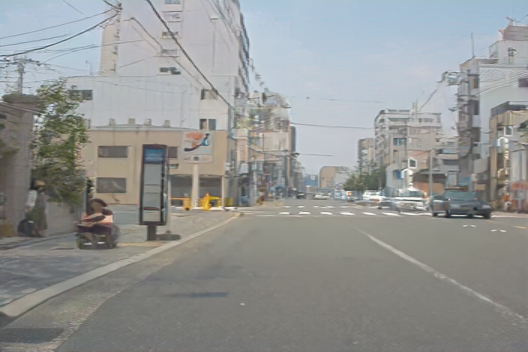}}
   \caption{InstaFormer} \end{subfigure}\hfill\\
   \vspace{-5pt}
\caption{\textbf{Qualitative comparison on INIT dataset~\cite{shen2019towards}:} (top to bottom) sunny$\rightarrow$night, night$\rightarrow$sunny, and cloudy$\rightarrow$sunny results.
Among the methods, ours preserves object details well and show realistic results.}
  \label{fig:qualitative}\vspace{-10pt}
\end{figure*}

\subsection{Experimental Results}
\paragraph{Qualitative Evaluation.}

\begin{figure} 
\centering
\vspace{10pt}
\includegraphics[width=1\linewidth]{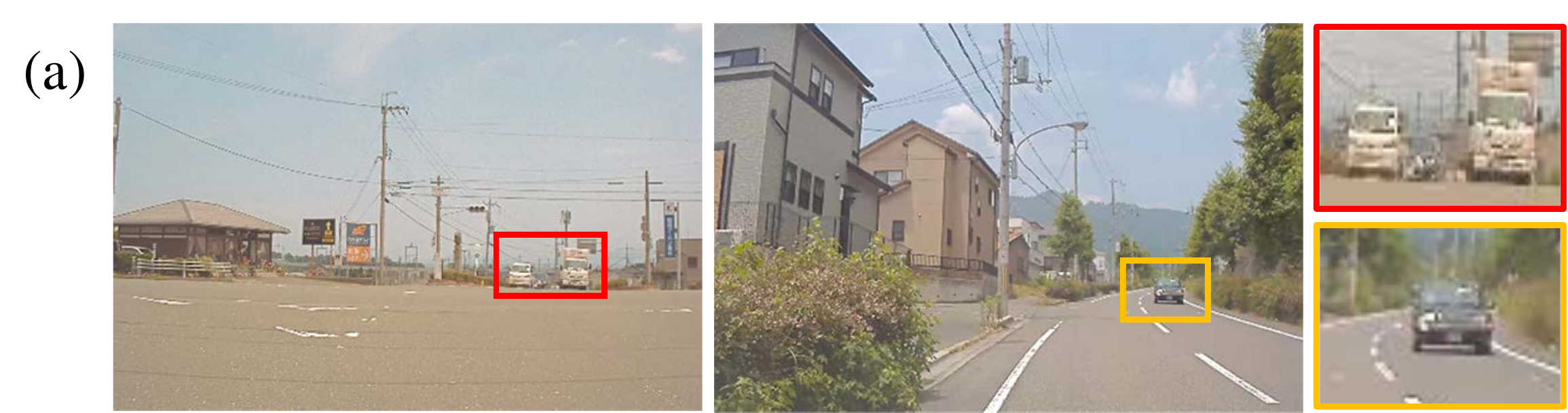}\\ \vspace{0.5pt}
\includegraphics[width=1\linewidth]{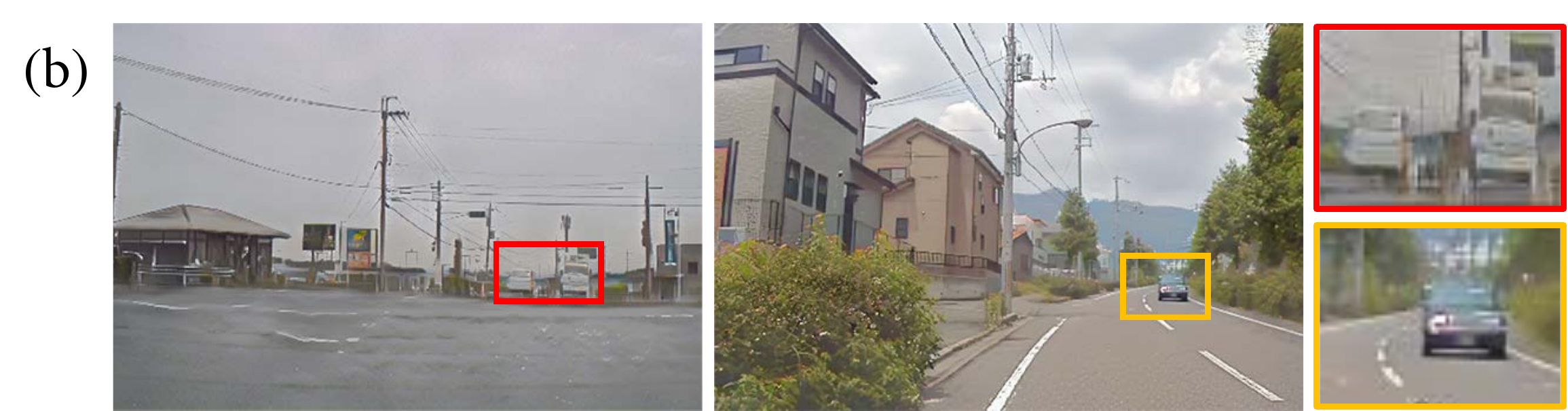}\\ \vspace{0.5pt}
\includegraphics[width=1\linewidth]{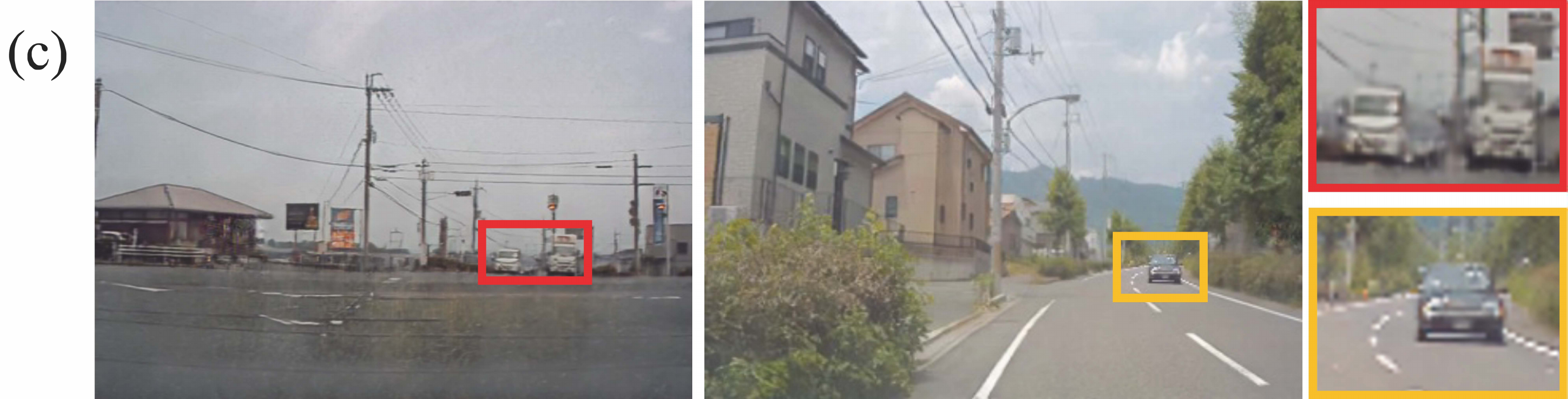}\\ \vspace{-4pt}
\caption{\textbf{Visual comparison with MGUIT~\cite{jeong2021memory}:}
(a) input, (b) MGUIT~\cite{jeong2021memory}, and (c) InstaFormer. We show the results for sunny$\rightarrow$rainy (left) and sunny$\rightarrow$cloudy (right).}
\label{fig:object} \vspace{-10pt}
\end{figure}

We first conduct qualitative comparisons of our method to CycleGAN~\cite{zhu2017unpaired}, UNIT~\cite{liu2017unsupervised}, MUNIT~\cite{huang2018multimodal}, DRIT~\cite{lee2018diverse}, and MGUIT~\cite{jeong2021memory} on sunny$\rightarrow$night, night$\rightarrow$sunny, sunny$\rightarrow$cloudy, and sunny$\rightarrow$rainy tasks in INIT dataset~\cite{shen2019towards}. As shown in~\figref{fig:qualitative}, our model generates higher quality translated results, particularly at object instance regions. Especially, as exemplified in the highlighted regions in~\figref{fig:object}, our model is good at capturing local regions within multiple instances thanks to Transformer-based architecture that simultaneously consider object instances and global image, and proposed instance-level contrastive learning. Note that our attention map visualization also proves this, well illustrated in~\figref{fig:attention}. 
Note that MGUIT~\cite{jeong2021memory} has access on their trained memory module during test-time, which is additional burden. \vspace{-10pt}

\paragraph{Quantitative Evaluation.}
Following the common practice~\cite{shen2019towards,bhattacharjee2020dunit,jeong2021memory}, we evaluate our InstaFormer with inception score (IS)~\cite{salimans2016improved} and conditional inception score (CIS)~\cite{huang2018multimodal}. Since the metrics above are related to diversity of translated images, we also evaluate our methods with fr\'{e}chet inception score (FID)~\cite{heusel2017gans} and structural similarity index measure (SSIM)~\cite{wang2004image} in terms of quality of translated images. Note that we evaluate the results under the same settings for all the methods. We adopt FID to measure the distance between distributions of real images and synthesized images in a deep feature domain. In addition, since SSIM index is an error measurement which is computed between the original content images and synthesized images, we apply to measure instance-wise structural consistency. It should be noted that for image translation tasks, there often exists some discrepancy between quantitative evaluations and human perceptions~\cite{borji2019pros}, thus the user study in the following would be a better precise metric.

As shown in~\tabref{tab1}, our InstaFormer outperforms the current state-of-the-art methods in terms of diversity (CIS, IS). Furthermore, in terms of global distribution, or instance-level similarity as shown in~\tabref{tab:fid}, FID and SSIM score show our InstaFormer tends to outperform prior methods in almost all the comparisons. In particular, results on SSIM demonstrate that our network is faithfully designed to encode an instance-awareness. 
Our method improves the FID score by a large margin compared to previous leading methods MGUIT~\cite{jeong2021memory} on INIT dataset~\cite{shen2019towards}.
\vspace{-10pt}

\begin{figure}[]
  \centering
  \vspace{-5pt}
  \begin{subfigure}{0.8\linewidth}{\includegraphics[width=1\linewidth]{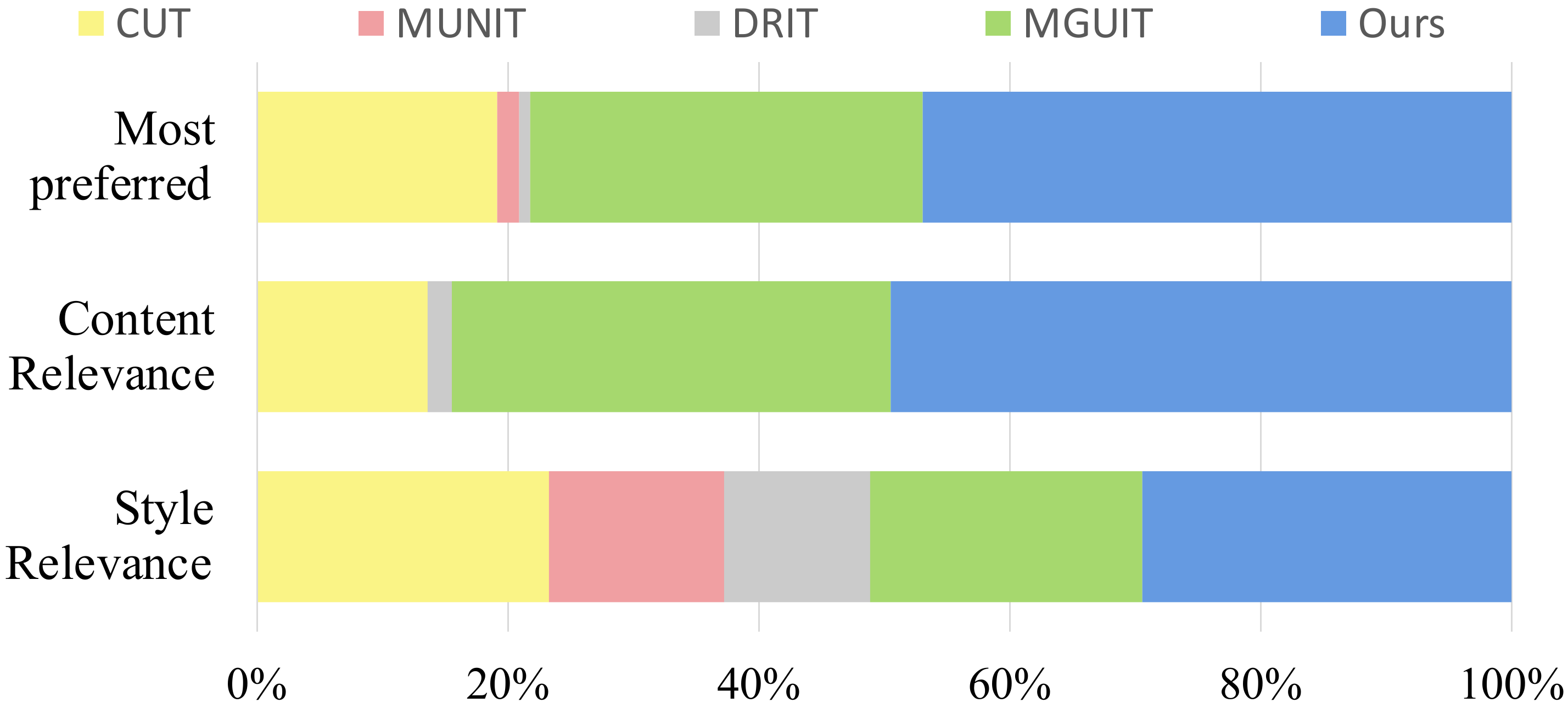}}
 \end{subfigure}\hfill\\
	\caption{\textbf{User study results on INIT dataset~\cite{shen2019towards}.} Our method is most preferred for overall quality, semantic consistency and style relevance, compared to CUT~\cite{park2020contrastive}, MUNIT~\cite{huang2018multimodal}, DRIT~\cite{lee2018diverse}, and MGUIT~\cite{jeong2021memory}.}
	\label{fig:userstudy_all}\vspace{-10pt}
\end{figure}

\paragraph{User Study.}
We also conducted a user study on 110 participants to evaluate the quality of synthesized images in the experiments with the following questions: ``Which do you think has better image quality in overall/ similar content to content image / represent style similar to target domain?" on INIT dataset, summarized in~\figref{fig:userstudy_all}. Our method ranks the first in every case, especially on content relevance and overall preference. Note that no standard evaluation metric has been emerged yet, human evaluation has an effect as evaluation metrics in image translation tasks.

\begin{table}[t]
\centering
\resizebox{1\linewidth}{!}{
\begin{tabular}{l|cc|cc|cc}
\hline
\multirow{2}{*}{Methods}	& \multicolumn{2}{c|}{sunny$\rightarrow$night} &	\multicolumn{2}{c|}{night$\rightarrow$sunny} & \multicolumn{2}{c}{Average}	
\\
& FID$\downarrow$ &SSIM$\uparrow$  & FID$\downarrow$&SSIM$\uparrow$  & FID$\downarrow$&SSIM$\uparrow$  \tabularnewline

\hline\hline
CUT~\cite{park2020contrastive}	&	\textbf{75.28} & 0.698	&	\underline{80.72}	& 0.634 & 	\textbf{78.00}&	0.666	\\
MUNIT~\cite{huang2018multimodal}	&	100.32	& 0.703 &	98.04	& 0.631 &	99.18& 0.680	\\
DRIT~\cite{lee2018diverse}	&	\underline{79.59}	& 0.312 &	99.33	& 0.266 &	89.46 & 0.289\\
MGUIT~\cite{jeong2021memory}	&	98.03	& \underline{0.836} &	82.17	& \textbf{0.848} &	90.10& \underline{0.842}	\\
\hline
InstaFormer & {84.72} & \textbf{0.872} & \textbf{71.65} &{\underline{0.818}}&{\underline{79.05}}&\textbf{0.845}\\
\hline
\end{tabular}}
\vspace{-5pt}
\caption{\textbf{Quantitative evaluation with FID~\cite{heusel2017gans} metric for data distribution and SSIM~\cite{wang2004image} index measured at each instance.}}
\label{tab:fid}
\end{table}

\begin{figure*}[t]
\centering
\begin{subfigure}{0.142\linewidth}{\includegraphics[width=1\linewidth]{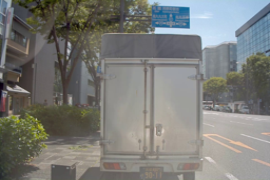}}
  \caption{Content image} \end{subfigure}\hfill	
\begin{subfigure}{0.142\linewidth}{\includegraphics[width=1\linewidth]{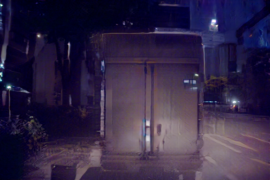}}
  \caption{InstaFormer} \end{subfigure}\hfill	
\begin{subfigure}{0.142\linewidth}{\includegraphics[width=1\linewidth]{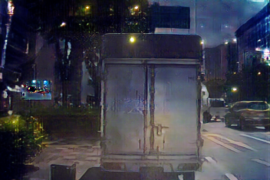}}
  \caption{MLP-Mixer~\cite{tolstikhin2021mlp}} \end{subfigure}\hfill
\begin{subfigure}{0.142\linewidth}{\includegraphics[width=1\linewidth]{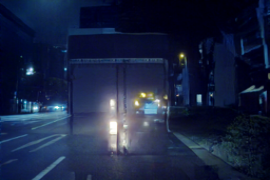}}
  \caption{w/o $\mathcal{L}^\mathrm{ins}_\mathrm{NCE}$} \end{subfigure}\hfill
\begin{subfigure}{0.142\linewidth}{\includegraphics[width=1\linewidth]{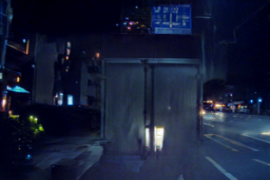}}
  \caption{w/o $\mathcal{L}^\mathrm{ins}_\mathrm{NCE}$, $\mathcal{T}$} \end{subfigure}\hfill
\begin{subfigure}{0.142\linewidth}{\includegraphics[width=1\linewidth]{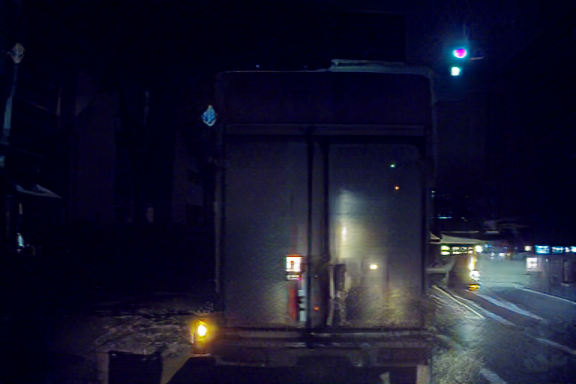}}
  \caption{CUT~\cite{park2020contrastive}} \end{subfigure}\hfill
\begin{subfigure}{0.142\linewidth}{\includegraphics[width=1\linewidth]{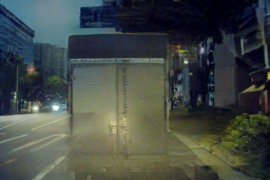}}
  \caption{w/o AdaIN} \end{subfigure}\hfill
\\
	\vspace{-5pt}
    \caption{\textbf{Ablation study on different settings:} instance-level loss ($\mathcal{L}^\mathrm{ins}_\mathrm{NCE}$), Transformer encoder ($\mathcal{T}$), normalization, and another backbone (MLP-Mixer). Note that CUT equals to the setting w/o $\mathcal{L}^\mathrm{ins}_\mathrm{NCE}$, $\mathcal{T}$, and AdaIN.}
	\label{fig:ablation} \vspace{-10pt} 
\end{figure*}

\begin{figure}[t]
  \centering
  \vspace{-5pt}
  \begin{subfigure}{0.8\linewidth}{\includegraphics[width=1\linewidth]{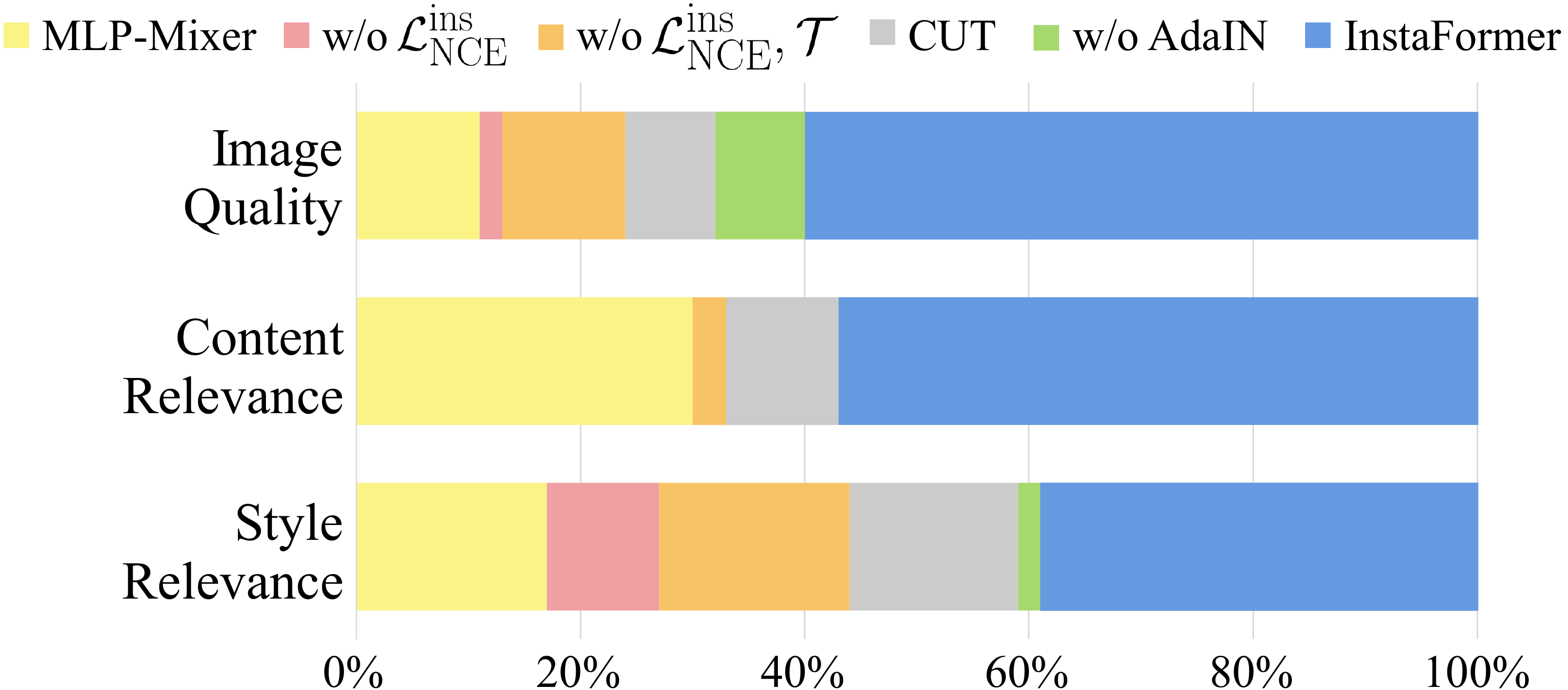}}
 \end{subfigure}\hfill\\
    \caption{\textbf{User study results on ablation study.}
    }
	\label{fig:userstudy_ablation}\vspace{-10pt}
\end{figure}

\subsection{Ablation Study}
In order to validate the effectiveness of each component in our method, we conduct a comprehensive ablation study. In particular, we analyze the effectiveness of instance-level loss ($\mathcal{L}^\mathrm{ins}_\mathrm{NCE}$), Transformer encoder ($\mathcal{T}$), and AdaIN, shown in~\figref{fig:ablation}. It should be noted that CUT~\cite{park2020contrastive} can be regarded as the setting without $\mathcal{L}^\mathrm{ins}_\mathrm{NCE}$, $\mathcal{T}$, AdaIN from InstaFormer.
Without $\mathcal{L}^\mathrm{ins}_\mathrm{NCE}$, our self-attention module has limited capability to focus on objects, thus generating images containing blurred objects, as evaluated in~\figref{fig:attention} as well. To validate the effect of $\mathcal{T}$ in our model, we conduct ablation experiments by replacing it with Resblocks (without $\mathcal{L}^\mathrm{ins}_\mathrm{NCE}$, $\mathcal{T}$). Without Transformers, it fails to capture global relationship between features. It is obvious that CUT~\cite{park2020contrastive} shows limited results containing artifacts, while InstaFormer dramatically improves object-awareness and quality of the generated image thanks to our architecture. Since AdaIN helps to understand global style by leveraging affine parameters, the result without AdaIN, which is replaced with LayerNorm, shows limited preservation on style with a single-modal output. 
We also validate our ablation study results on human evaluation. 110 participants are asked to consider three aspects: overall quality, semantic consistency and style consistency, summarized in \figref{fig:userstudy_ablation}, where we also validate the superiority of each proposed component.

In addition, we conduct experiments using MLP-Mixer~\cite{tolstikhin2021mlp}-based aggregator that replaces $\mathcal{T}$ consisted of ViT~\cite{dosovitskiy2020image} blocks to justify robustness of our framework.
\figref{fig:ablation}(c) shows result examples by MLP-Mixer~\cite{tolstikhin2021mlp}-based aggregator. Although ViT-based model is slightly better on MLP-Mixer~\cite{tolstikhin2021mlp}-based model in overall quality in~\figref{fig:ablation}(b), the object instance and style representation are faithfully preserved, which indicates that our method can be adopted in another Transformer backbone. 

\subsection{Domain Adaptive Object Detection}
Additionally, we evaluate our method on the task of unsupervised domain adaptation for object detection.
We follow the experimental setup in DUNIT~\cite{bhattacharjee2020dunit}. We used Faster-RCNN~\cite{ren2015faster} as baseline detector. In~\tabref{tab:domain}, we report the per-class average precisions (AP) for the KITTI→Cityscapes case~\cite{geiger2012we,cordts2016cityscapes}. Compared to DUNIT~\cite{bhattacharjee2020dunit} and MGUIT~\cite{jeong2021memory}, our model shows impressive results. It should be noted that we do not access any information about bounding box information on test-time, while DUNIT contains object detection network and MGUIT has access to trained external memory by reading class-aware features.
In particular, our model significantly outperforms other methods in almost all classes, which indicates that our suggested instance loss has strength on instance-awareness.

\begin{table}[]
\centering
\resizebox{0.75\linewidth}{!}{
\begin{tabular}{l|c|c|c|c|c}
\hline
Method	&	Pers	&	Car	&	Truc. &Bic & mAP		\\ 
\hline\hline
DT~\cite{inoue2018cross}	&	28.5	&	40.7	&	25.9	&	29.7	&	31.2    \\
DAF~\cite{huang2018multimodal}	&	39.2	&	40.2	&	25.7	&	48.9	&	38.5    \\
DARL~\cite{kim2019diversify}	&	46.4	&	58.7	&	27.0	&	49.1    &	45.3	\\
DAOD~\cite{rodriguez2019domain}	&	47.3	&	59.1	&	28.3	&	49.6    &	46.1    \\
DUNIT~\cite{bhattacharjee2020dunit}	&	\underline{60.7}	&	65.1	&	32.7	&	\underline{57.7}    &	54.1	\\
MGUIT~\cite{jeong2021memory}&	58.3	&\underline{68.2}	&	\underline{33.4}	& \textbf{58.4}	&	\underline{54.6}\\
\hline
InstaFormer & \textbf{61.8} & \textbf{69.5} & \textbf{35.3} & 55.3 & \textbf{55.5}\\
\hline
\end{tabular}} 
\vspace{-5pt}
\caption{\textbf{Results for domain adaptive detection.}
We compare the per-class Average Precision for KITTI $\rightarrow$ CityScape.}
\label{tab:domain}\vspace{-10pt}
\end{table}

\section{Conclusion}
In this paper, we have presented Transformer-based networks, dubbed InstaFormer, for instance-aware image-to-image translation, which enables boosting the translation of object instances as well as global image. By simultaneously considering instance-level features and global-level features with Transformers, we learned an interaction between not only object instance and global image, but also different instances. To improve the instance-awareness during translation, we proposed an instance-level content contrastive loss. Experiments on various datasets with evaluation metrics have shown that our framework outperforms the existing solutions for instance-aware I2I.

\noindent\textbf{Acknowledgements.}
This research was supported by the MSIT, Korea (IITP-2022-2020-0-01819, ICT Creative Consilience program), and National Research Foundation of Korea (NRF-2021R1C1C1006897). 

{\small
\bibliographystyle{ieee_fullname}
\bibliography{egbib.bib}
}

\newpage
\onecolumn

\appendix


\pagenumbering{gobble}
In this appendix, we describe detailed network architecture, PyTorch-like pseudo-code, and additional results for ``InstaFormer: Instance-Aware Image-to-Image Translation with Transformer".

\section{Experimental Details}
\subsection{Network Architecture of InstaFormer} 

We first summarize the detailed network architecture of our InstaFormer in \tabref{tab:1}.
We basically follow the content encoder and generator architecture from CUT\cite{park2020contrastive} with Transformer blocks and style encoder. The double-line inside Encoder table indicates the end of content encoder, while style encoder has same structure with content encoder but has additional Adaptive Avg. Pool. and Conv-4 as shown below.
$(.)$ in the convolution indicates the padding technique, and the structures of Upsample , Downsample and discriminator are the same as those of CUT\cite{park2020contrastive}.

	\begin{table}[h] 
	\centering
	 \scalebox{0.9}{
		
			\begin{tabular}{>{\centering}m{0.18\linewidth} >{\centering}m{0.30\linewidth} 
					>{\centering}m{0.30\linewidth}}
					&\textbf{Encoder}& \tabularnewline
				\hlinewd{0.8pt}
				Layer & Parameters $(\mathtt{in},\mathtt{out},\mathtt{k},\mathtt{s},\mathtt{p})$ & Output shape $(C \times H \times W)$ \tabularnewline
				\hline
				Conv-1 (Reflection) & $(3,64,7,1,3)$ & $(64,352,352)$  \tabularnewline
				InstanceNorm & - & $(64,352,352)$  \tabularnewline		
				ReLU & - & $(64,352,352)$  \tabularnewline		\hline
				Conv-2 (Zeros) & $(64,128,3,1,1)$ & $(128,352,352)$ \tabularnewline
				InstanceNorm & - & $(128,352,352)$  \tabularnewline		
				ReLU & - & $(128,352,352)$  \tabularnewline		\hline
				Downsample & - & $(128,176,176)$ \tabularnewline		
				Conv-3 (Zeros) & $(128,256,3,1,1)$ & $(256,176,176)$ \tabularnewline
				InstanceNorm & - & $(256,176,176)$  \tabularnewline		
				ReLU & - & $(256,176,176)$  \tabularnewline		
				DownSample & - & $(256,88,88)$ \tabularnewline \hline\hline
				AdaptiveAvgPool & - & $(256,1,1)$ \tabularnewline
				Conv-4 & $(256,8,1,1,0)$ & $(8,1,1)$ \tabularnewline
				\hlinewd{0.8pt}
                &&\tabularnewline 
                &\textbf{Transformer Aggregator}& \tabularnewline \hlinewd{0.8pt}
				Layer & Parameters $(\mathtt{in},\mathtt{out})$ & Output shape $(C)$ \tabularnewline
				\hline
				AdaptiveInstanceNorm & - & $(1024)$ \tabularnewline
                Linear-1 & $(1024,3072)$ & $(3072)$ \tabularnewline
				Attention & - & $(1024)$ \tabularnewline
				Linear-2 & $(1024,1024)$ & $(1024)$ \tabularnewline
				AdaptiveInstanceNorm & - & $(1024)$ \tabularnewline
                Linear-3 & $(1024,4096)$ & $(4096)$ \tabularnewline
				GELU & - & $(4096)$ \tabularnewline
				Linear-4 & $(4096,1024)$ & $(1024)$ \tabularnewline

				\hlinewd{0.8pt}
                &&\tabularnewline 
                &\textbf{Generator}& \tabularnewline \hlinewd{0.8pt}
				Layer & Parameters $(\mathtt{in},\mathtt{out},\mathtt{k},\mathtt{s},\mathtt{p})$ & Output shape $(C \times H \times W)$ \tabularnewline
				\hline
                UpSample & - & $(256,176,176)$ \tabularnewline 
                Conv-1 (Zeros) & $(256,128,3,1,1)$ & $(128,176,176)$  \tabularnewline
				InstanceNorm & - & $(128,176,176)$  \tabularnewline		
				ReLU & - & $(128,176,176)$  \tabularnewline		\hline
                UpSample & - & $(128,352,352)$ \tabularnewline 
                Conv-2 (Zeros) & $(128,64,3,1,1)$ & $(64,352,352)$  \tabularnewline
				InstanceNorm & - & $(64,352,352)$  \tabularnewline		
				ReLU & - & $(64,352,352)$  \tabularnewline		\hline
				Conv-3 (ReflectionPad) & $(64,3,7,1,3)$ & $(3,352,352)$ \tabularnewline
				Tanh & - & $(3,352,352)$  \tabularnewline	
				\hline
				\hlinewd{0.8pt}
			\end{tabular}
		}
		\caption{\textbf{Network architecture of our InstaFormer.}}
		\label{tab:1}
	\end{table}

\newpage
\subsection{PyTorch-like Pseudo-code}
\paragraph{Pseudo-code for Instance-level Content Loss.}
Here we provide the (PyTorch-like) pseudo-code for $\mathcal{L}^\mathrm{ins}_\mathrm{NCE}$ in InstaFormer. 
To re-emphasize, our simple instance-level content loss learns the representations for a translation task, effectively focusing on local object-region; this is unlike classical PatchNCE loss that utilizes the regular-grid patches from features in unconditional way.

\vspace{-10pt}
\begin{figure*}[h]
    \centering
    \begin{center}
	{\includegraphics[width=1\linewidth]{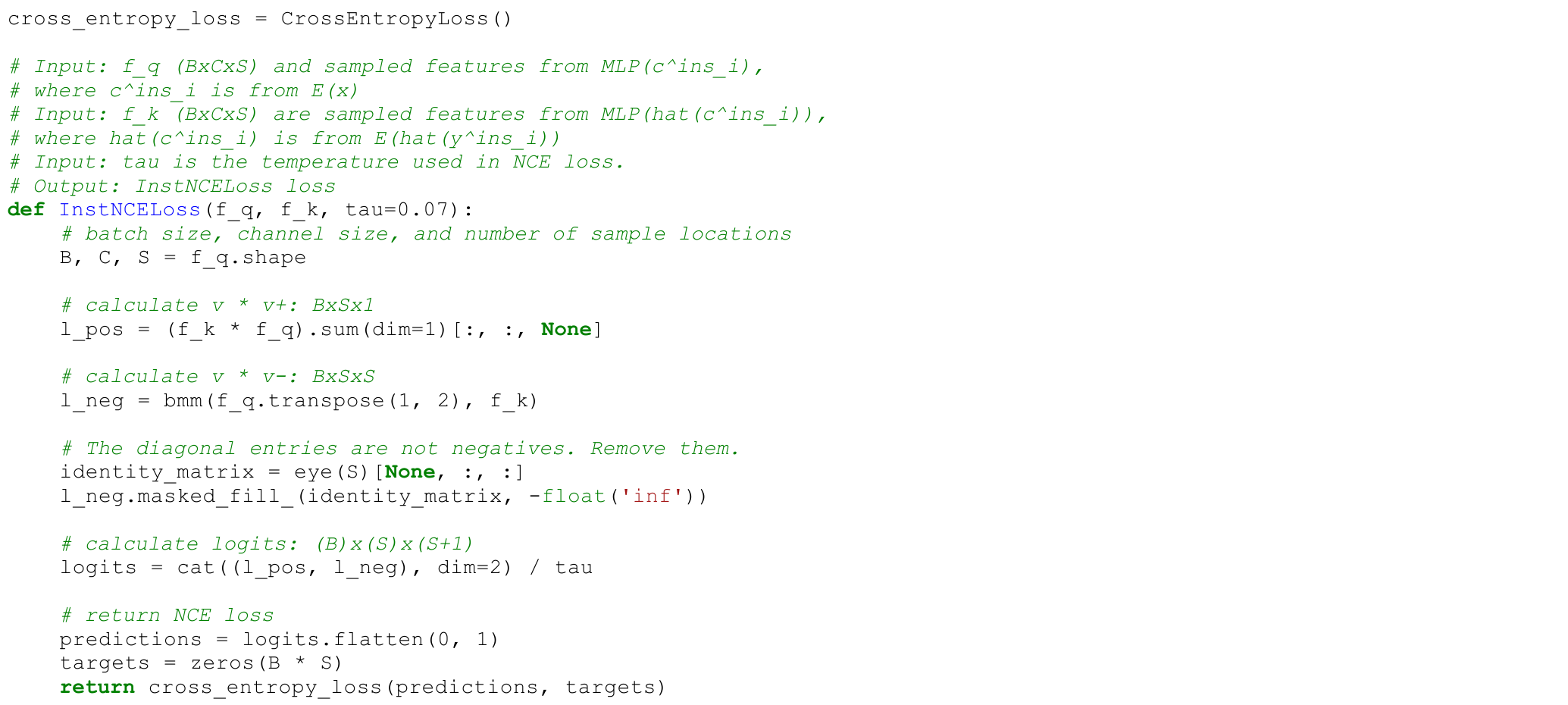}}\hfill
	\end{center}
	\label{fig:code1}\vspace{-10pt}
\end{figure*}

\paragraph{Pseudo-code for Transformer Aggregator.} We also provide pseudo-code for the input of $\mathcal{T}$, in order to show how our novel technique aggregates \emph{instance}-level content features and \emph{global}-level content features simultaneously. This enables the model to pay more attention to the relationships between global scenes and object instances.

\vspace{-10pt}
\begin{figure*}[h]
    \centering
    \begin{center}
	{\includegraphics[width=1\linewidth]{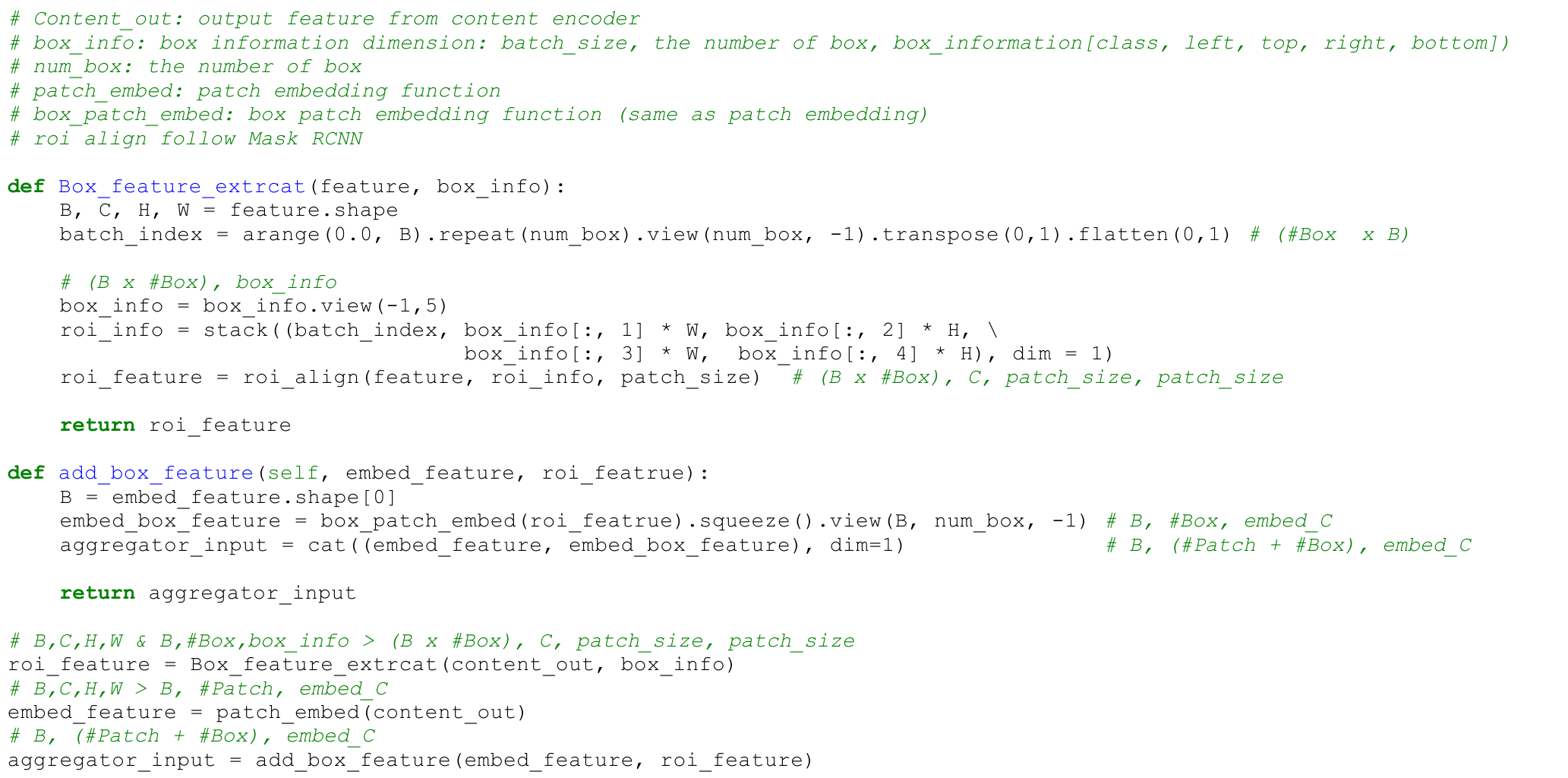}}\hfill
	\end{center}
	\label{fig:code2}\vspace{-10pt}
\end{figure*}

\newpage
\section{Additional Results}
\subsection{Visualization of Multi-modal Image Translation} 
\begin{figure*}[h]
\centering
\begin{subfigure}
{0.247\linewidth}\includegraphics[width=1\linewidth]{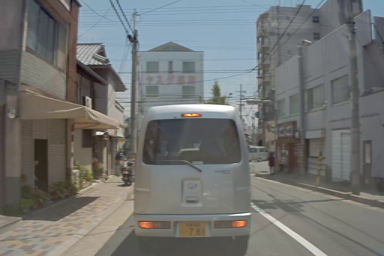}
\end{subfigure}\hfill
\begin{subfigure} 
{0.247\linewidth}\includegraphics[width=1\linewidth]{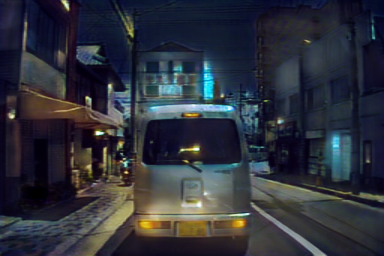}\end{subfigure}\hfill
\begin{subfigure} 
{0.247\linewidth}\includegraphics[width=1\linewidth]{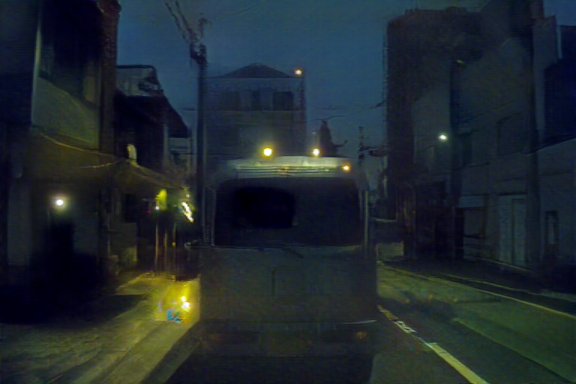}\end{subfigure}\hfill
\begin{subfigure} 
{0.247\linewidth}\includegraphics[width=1\linewidth]{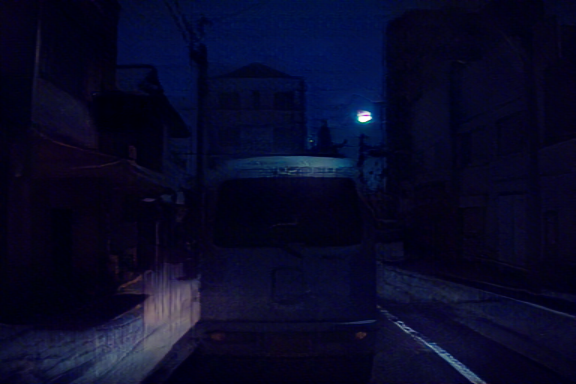}\end{subfigure}\hfill \\
\begin{subfigure} 
{0.247\linewidth}\includegraphics[width=1\linewidth]{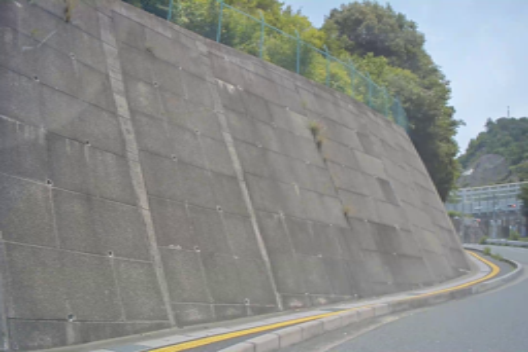}  \caption{Input image (Sunny)}\end{subfigure}\hfill
\begin{subfigure} 
{0.247\linewidth}\includegraphics[width=1\linewidth]{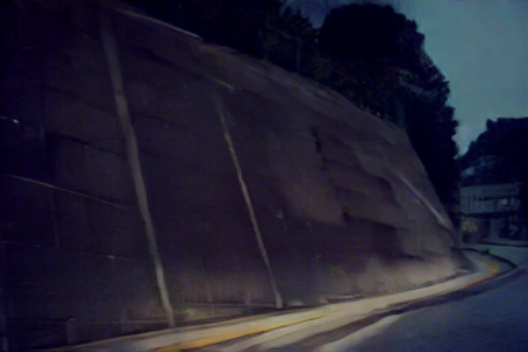}\caption{Translated image 1 (Night)}\end{subfigure}\hfill
\begin{subfigure} 
{0.247\linewidth}\includegraphics[width=1\linewidth]{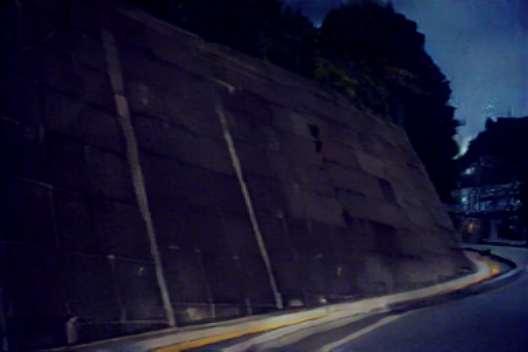}\caption{Translated image 2 (Night)}\end{subfigure}\hfill
\begin{subfigure} 
{0.247\linewidth}\includegraphics[width=1\linewidth]{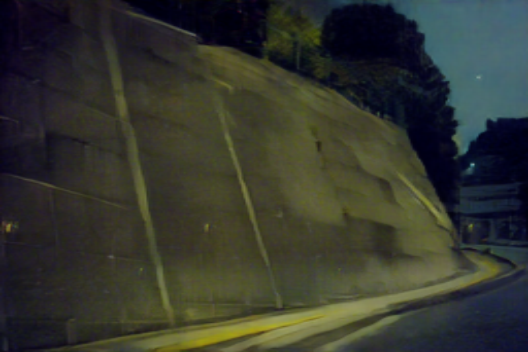}\caption{Translated image 3 (Night)}\end{subfigure}\hfill \\

\vspace{-5pt}
\caption{\textbf{Results of multi-modal image translation.}
We use randomly sampled style codes to generate night images from a sunny image.}
\vspace{-10pt}
\label{fig:multi}
\end{figure*}

We visualize the multimodal translated results in~\figref{fig:multi}. Our InstaFormer generates not only high-quality visual results, but also produces results with large diversity.

\subsection{Qualitative Results of Domain Adaptive Object Detection} 
\begin{figure*}[h]
  \centering
   \begin{subfigure}{0.498\linewidth}{\includegraphics[width=1\linewidth]{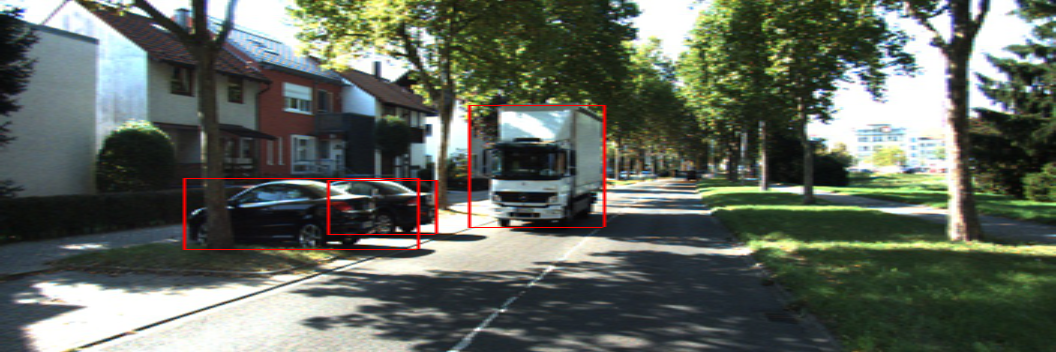}} \end{subfigure}\hfill
  \begin{subfigure}{0.498\linewidth}{\includegraphics[width=1\linewidth]{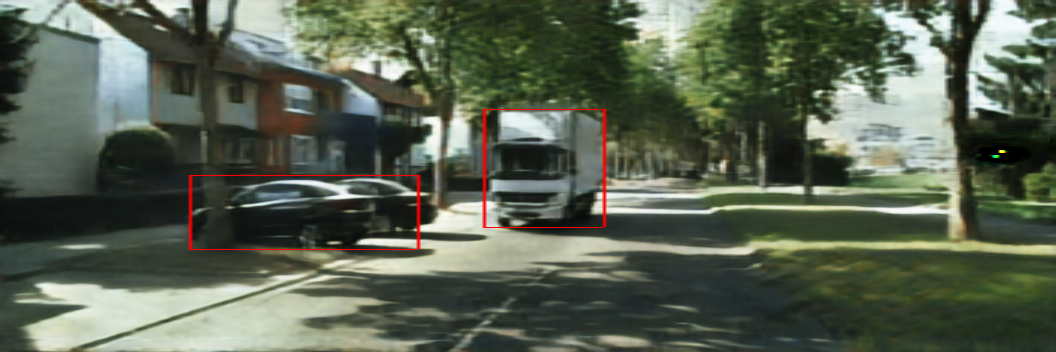}} \end{subfigure}\hfill \\
   \begin{subfigure}{0.498\linewidth}{\includegraphics[width=1\linewidth]{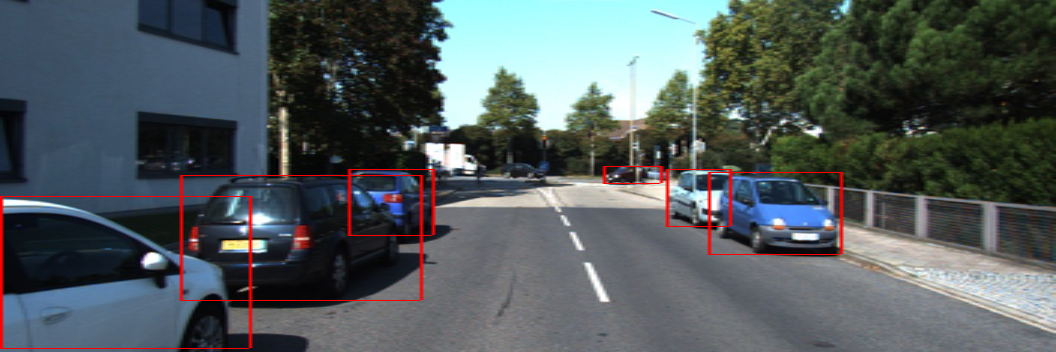}}  \caption{Input image (KITTI)}\end{subfigure}\hfill
  \begin{subfigure}{0.498\linewidth}{\includegraphics[width=1\linewidth]{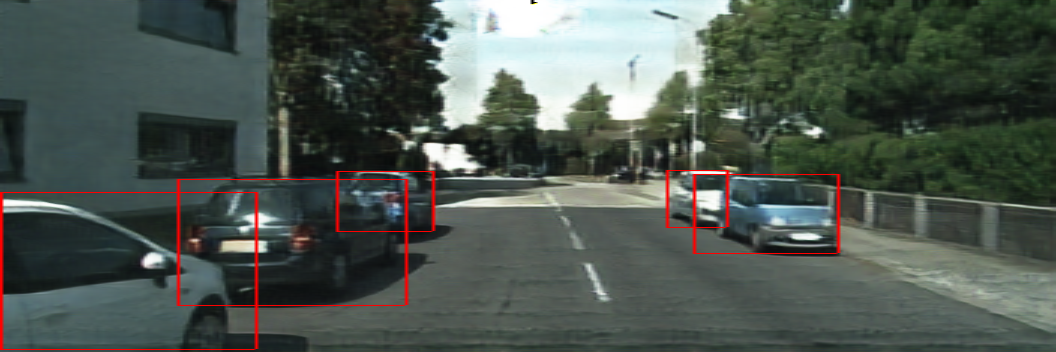}}   \caption{Translated image (CityScape)}\end{subfigure}\hfill\\

   \vspace{-5pt}
\caption{\textbf{Qualitative results for domain adaptive detection for KITTI $\rightarrow$ CityScape.}}

  \label{fig:detection}\vspace{-10pt}
\end{figure*}

In the main paper, we have evaluated our method on the task of unsupervised domain adaptation for object detection providing quantitative results. In this section, we also show the qualitative results of the task in~\figref{fig:detection}. Our model successfully works on complex domain adaptation tasks.

\newpage
\subsection{Comparison with the State-of-the-Art I2I methods.}
\begin{figure}[h]
\centering
\begin{subfigure} 
{0.195\linewidth}\includegraphics[width=1\linewidth]{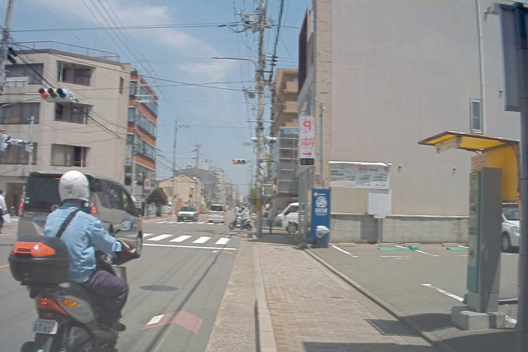}  \end{subfigure}\hfill
\begin{subfigure} 
{0.195\linewidth}\includegraphics[width=1\linewidth]{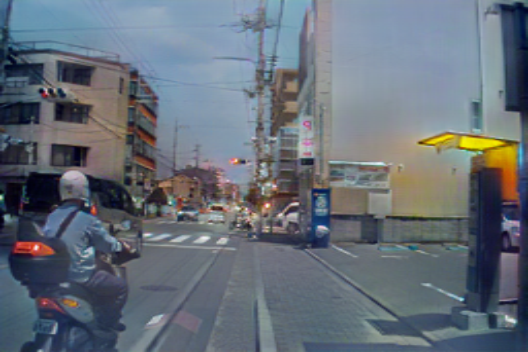}\end{subfigure}\hfill
\begin{subfigure} 
{0.195\linewidth}\includegraphics[width=1\linewidth]{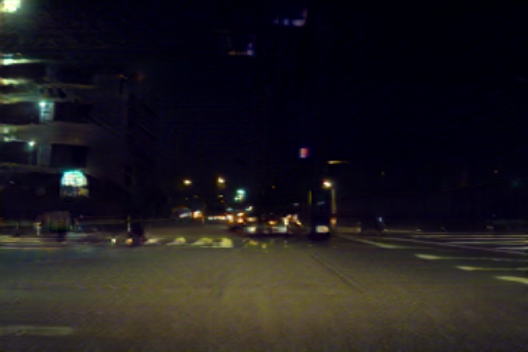}\end{subfigure}\hfill
\begin{subfigure} 
{0.195\linewidth}\includegraphics[width=1\linewidth]{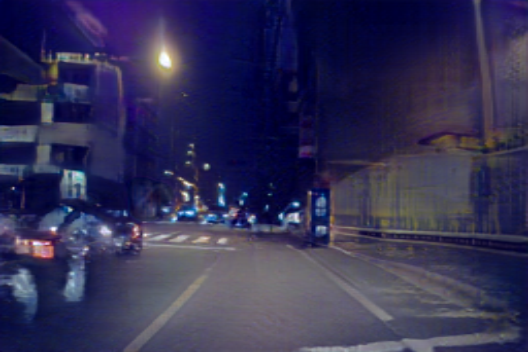}\end{subfigure}\hfill 
\begin{subfigure} 
{0.195\linewidth}\includegraphics[width=1\linewidth]{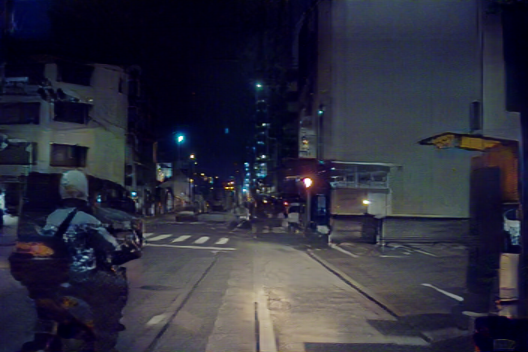}\end{subfigure}\hfill \\

\begin{subfigure} 
{0.195\linewidth}\includegraphics[width=1\linewidth]{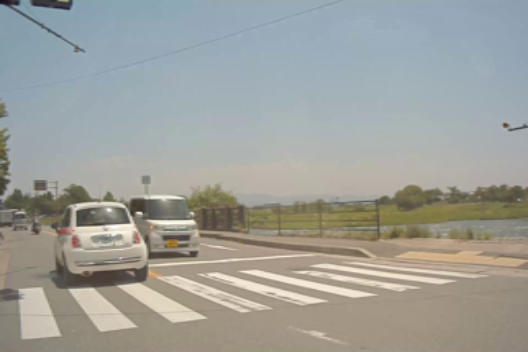}  \caption{Input}\end{subfigure}\hfill
\begin{subfigure} 
{0.195\linewidth}\includegraphics[width=1\linewidth]{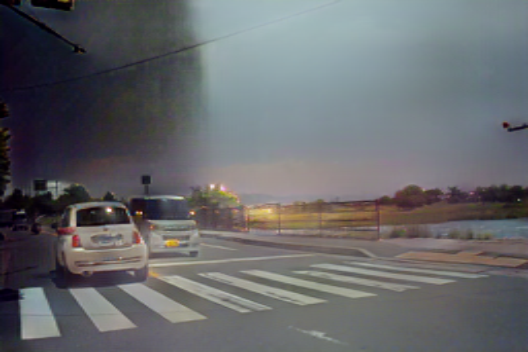}\caption{TSIT}\end{subfigure}\hfill
\begin{subfigure} 
{0.195\linewidth}\includegraphics[width=1\linewidth]{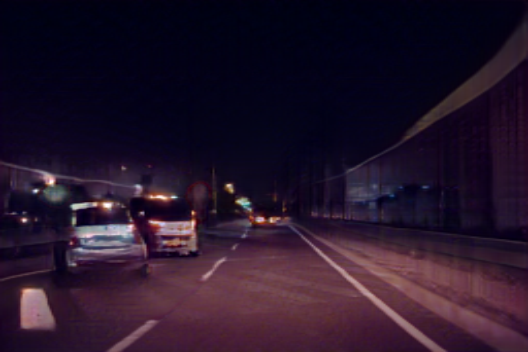}\caption{StarGAN2}\end{subfigure}\hfill
\begin{subfigure} 
{0.195\linewidth}\includegraphics[width=1\linewidth]{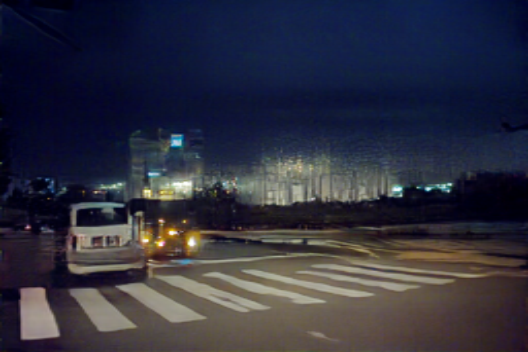}\caption{Smoothing}\end{subfigure}\hfill 
\begin{subfigure} 
{0.195\linewidth}\includegraphics[width=1\linewidth]{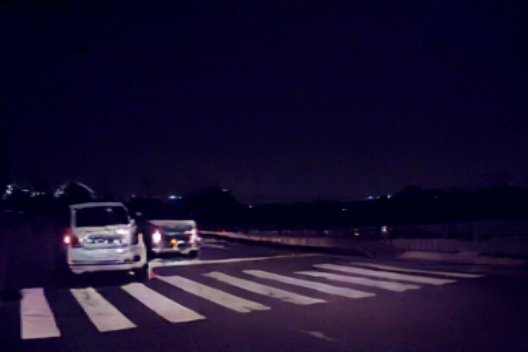}\caption{Ours}\end{subfigure}\hfill \\
	\vspace{-5pt}
    \caption{\textbf{Comparison with TSIT~\cite{jiang2020tsit}, StarGANv2~\cite{choi2020stargan} and Smoothing~\cite{liu2021smoothing}.}}
	\label{fig:moreresults}\vspace{-5pt}
\end{figure}

As shown in \figref{fig:moreresults} and \tabref{tab:tsit}, we additionally compare InstaFormer with SoTA I2I methods, such as TSIT~\cite{jiang2020tsit}, StarGANv2~\cite{choi2020stargan} and Smoothing~\cite{liu2021smoothing} on INIT dataset~\cite{shen2019towards} for sunny$\rightarrow$night.

\begin{table}[h]
\centering
\resizebox{0.25\linewidth}{!}{
\begin{tabular}{l|cc}
\hline
\multirow{2}{*}{Methods}	& \multicolumn{2}{c}{sunny$\rightarrow$night} 
\\
& FID$\downarrow$ &SSIM$\uparrow$  \tabularnewline

\hline\hline
TSIT	&	90.28 & 0.822 \\
StarGANv2	&	88.49	& 0.545 \\
Smoothing & 85.28 & 0.667 \\
\hline
InstaFormer &  \textbf{ 84.72 } & \textbf{ 0.872 } \\
\hline

\end{tabular}}
\vspace{-5pt}
\caption{\textbf{More quantitative evaluation.}}
\label{tab:tsit}\vspace{-10pt}
\end{table}

\subsection{Additional Examples on Ablation Study} 
\begin{figure*}[h]
\centering

\begin{subfigure}{0.142\linewidth}{\includegraphics[width=1\linewidth]{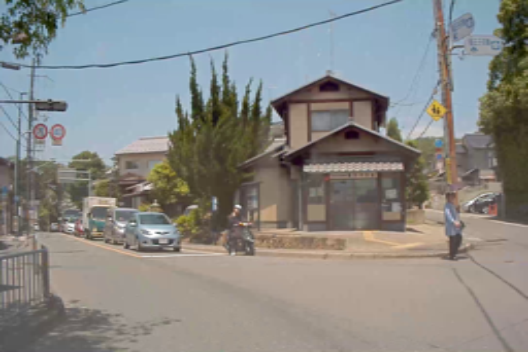}}
  \caption{Content image} \end{subfigure}\hfill	
\begin{subfigure}{0.142\linewidth}{\includegraphics[width=1\linewidth]{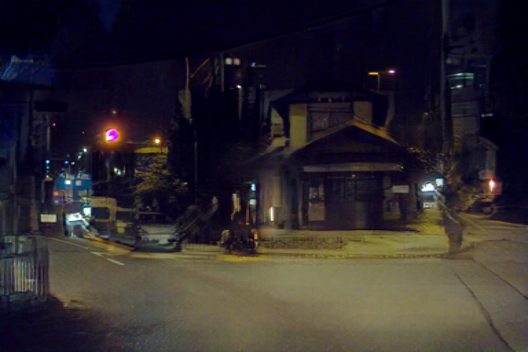}}
  \caption{InstaFormer} \end{subfigure}\hfill	
\begin{subfigure}{0.142\linewidth}{\includegraphics[width=1\linewidth]{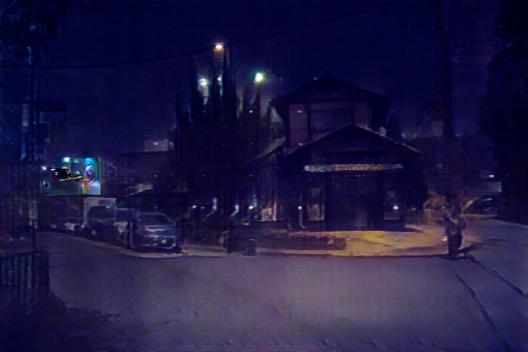}}
  \caption{MLP-Mixer~\cite{tolstikhin2021mlp}} \end{subfigure}\hfill
\begin{subfigure}{0.142\linewidth}{\includegraphics[width=1\linewidth]{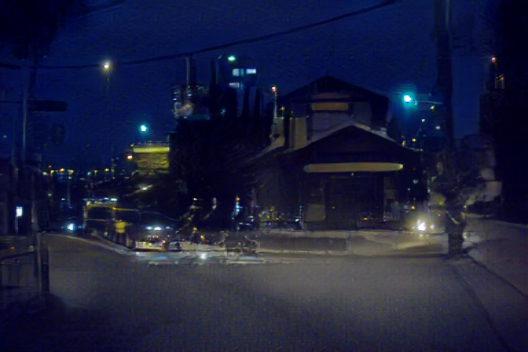}}
  \caption{w/o $\mathcal{L}^\mathrm{ins}_\mathrm{NCE}$} \end{subfigure}\hfill
\begin{subfigure}{0.142\linewidth}{\includegraphics[width=1\linewidth]{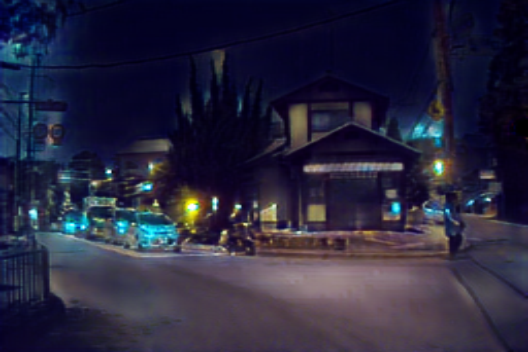}}
  \caption{w/o $\mathcal{L}^\mathrm{ins}_\mathrm{NCE}$, $\mathcal{T}$} \end{subfigure}\hfill
\begin{subfigure}{0.142\linewidth}{\includegraphics[width=1\linewidth]{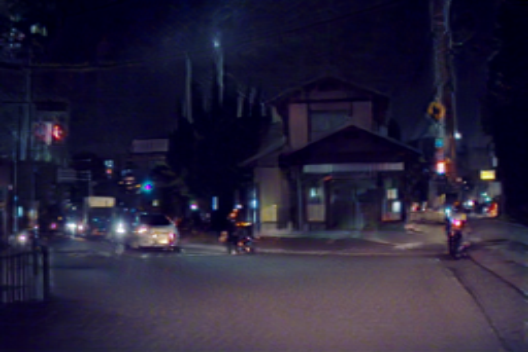}}
  \caption{CUT~\cite{park2020contrastive}} \end{subfigure}\hfill
\begin{subfigure}{0.142\linewidth}{\includegraphics[width=1\linewidth]{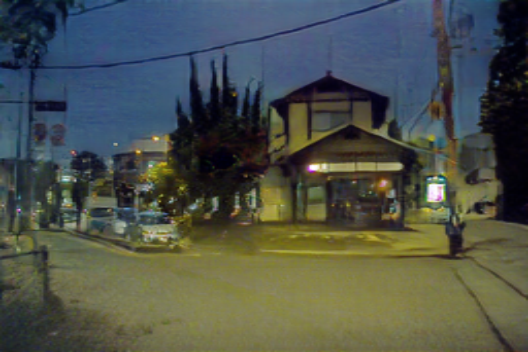}}
  \caption{w/o AdaIN} \end{subfigure}\hfill
\\
	\vspace{-5pt}
    \caption{\textbf{Ablation study on different settings:} instance-level loss ($\mathcal{L}^\mathrm{ins}_\mathrm{NCE}$), Transformer encoder ($\mathcal{T}$), normalization, and another backbone (MLP-Mixer). Note that CUT equals to the setting w/o $\mathcal{L}^\mathrm{ins}_\mathrm{NCE}$, $\mathcal{T}$, and AdaIN.}
	\label{fig:supp_abl}  
\end{figure*}

In the main paper, we have examined the impacts of instance-level loss ($\mathcal{L}^\mathrm{ins}_\mathrm{NCE}$), Transformer encoder ($\mathcal{T}$), normalization, and another backbone (MLP-Mixer). CUT equals to the setting w/o $\mathcal{L}^\mathrm{ins}_\mathrm{NCE}$, $\mathcal{T}$, and AdaIN.
We provide more examples on INIT dataset~\cite{shen2019towards}, depicted in~\figref{fig:supp_abl}. Our InstaFormer produces better visual results. In particular, as shown in~\figref{fig:supp_abl} (d), tiny objects tend to disappear or be blurred without $\mathcal{L}^\mathrm{ins}_\mathrm{NCE}$.

\newpage
\subsection{Additional Translation Results} 
\begin{figure*}[h]
  \centering
  \begin{subfigure}{0.247\linewidth}{\includegraphics[width=1\linewidth]{figure/fig2-3_2.png}}
  \end{subfigure}\hfill
  \begin{subfigure}{0.247\linewidth}{\includegraphics[width=1\linewidth]{figure/fig2-3_1.png}}
  \end{subfigure}\hfill
  \begin{subfigure}{0.247\linewidth}{\includegraphics[width=1\linewidth]{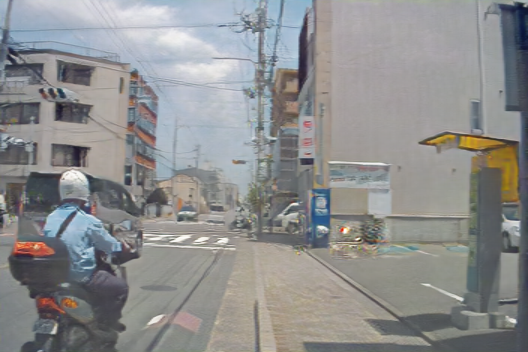}}
  \end{subfigure}\hfill
  \begin{subfigure}{0.247\linewidth}{\includegraphics[width=1\linewidth]{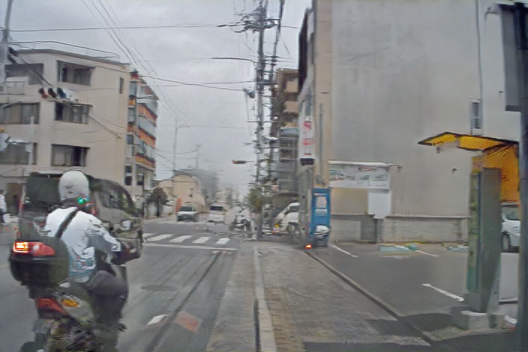}}
  \end{subfigure}\hfill\\
  
  \begin{subfigure}{0.247\linewidth}{\includegraphics[width=1\linewidth]{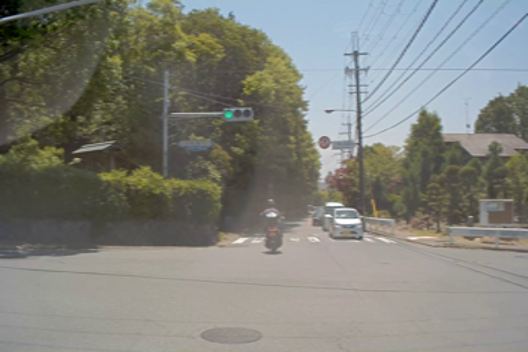}}
  \end{subfigure}\hfill
  \begin{subfigure}{0.247\linewidth}{\includegraphics[width=1\linewidth]{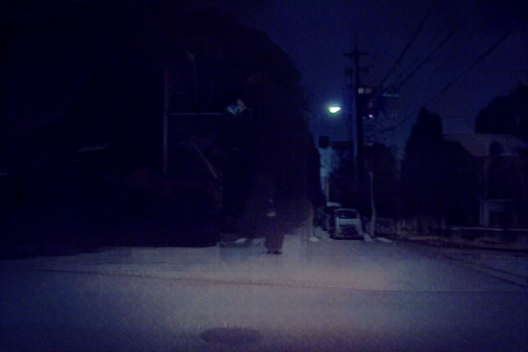}}
  \end{subfigure}\hfill
  \begin{subfigure}{0.247\linewidth}{\includegraphics[width=1\linewidth]{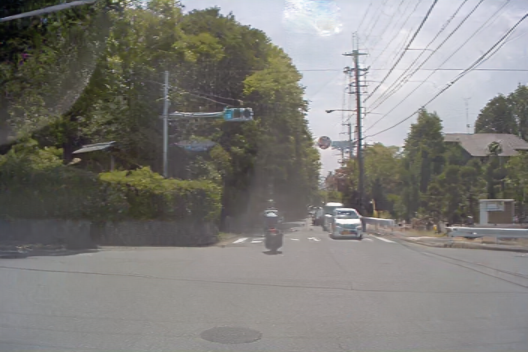}}
  \end{subfigure}\hfill
  \begin{subfigure}{0.247\linewidth}{\includegraphics[width=1\linewidth]{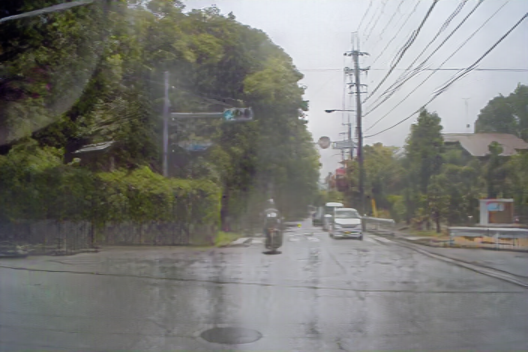}}
  \end{subfigure}\hfill\\
 
  \begin{subfigure}{0.247\linewidth}{\includegraphics[width=1\linewidth]{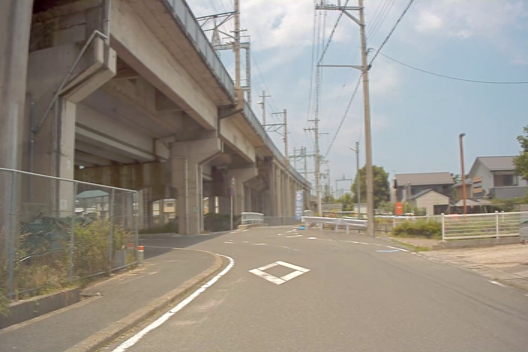}}
\end{subfigure}\hfill
  \begin{subfigure}{0.247\linewidth}{\includegraphics[width=1\linewidth]{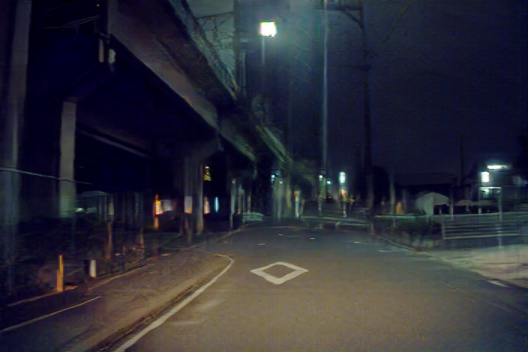}}
 \end{subfigure}\hfill 
  \begin{subfigure}{0.247\linewidth}{\includegraphics[width=1\linewidth]{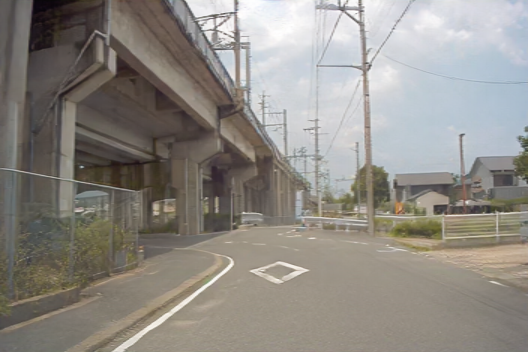}}
 \end{subfigure}\hfill
  \begin{subfigure}{0.247\linewidth}{\includegraphics[width=1\linewidth]{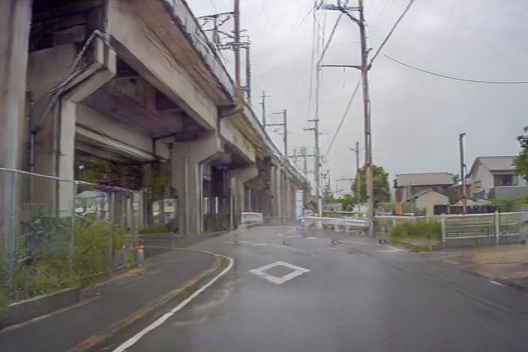}}
 \end{subfigure}\hfill\\
 
   \begin{subfigure}{0.247\linewidth}{\includegraphics[width=1\linewidth]{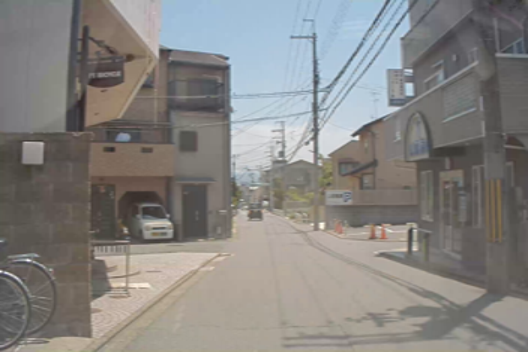}}
  \caption{Input image (Sunny)}\end{subfigure}\hfill
  \begin{subfigure}{0.247\linewidth}{\includegraphics[width=1\linewidth]{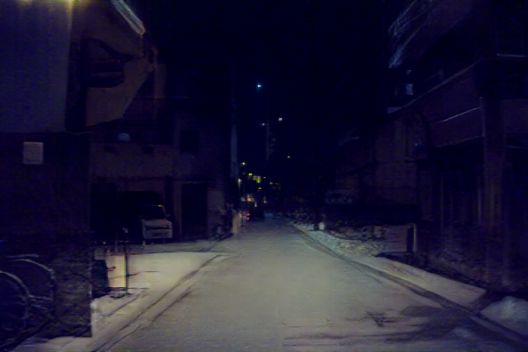}}
   \caption{Translated image (Night)}\end{subfigure}\hfill
  \begin{subfigure}{0.247\linewidth}{\includegraphics[width=1\linewidth]{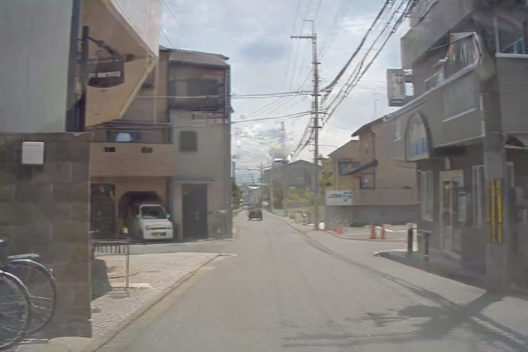}}
   \caption{Translated image (Cloudy)}\end{subfigure}\hfill
  \begin{subfigure}{0.247\linewidth}{\includegraphics[width=1\linewidth]{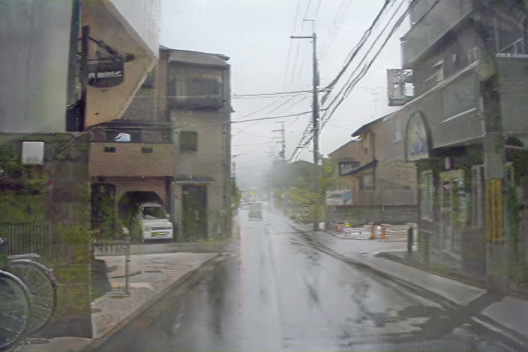}}
   \caption{Translated image (Rainy)}\end{subfigure}\hfill\\

   \vspace{-5pt}
\caption{\textbf{Qualitative comparison on INIT dataset~\cite{shen2019towards}:} (left to right) sunny, sunny$\rightarrow$night, sunny$\rightarrow$cloudy, and sunny$\rightarrow$rainy results. Our method achieves high-quality of realistic results while preserving object details  as well.}

  \label{fig:qual_supp}\vspace{-10pt}
\end{figure*}

In the main paper, we have shown some of our results of instance-aware image-to-image translation. Here, we show additional results in \figref{fig:qual_supp} to demonstrate the robustness of InstaFormer.

\newpage
\subsection{Variants of InstaFormer Architecture}
\begin{figure*}[h]
  \begin{subfigure}{0.247\linewidth}
  {\includegraphics[width=1\linewidth]{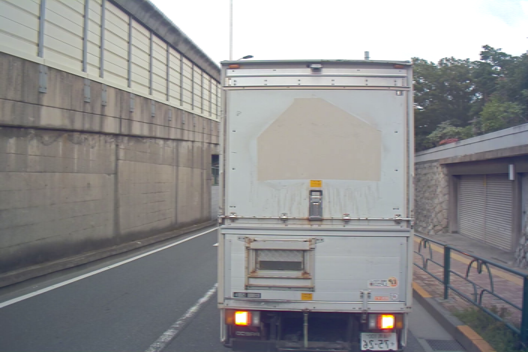}}
  \caption{Input image}\end{subfigure}\hfill
  \begin{subfigure}{0.247\linewidth}
   {\includegraphics[width=1\linewidth]{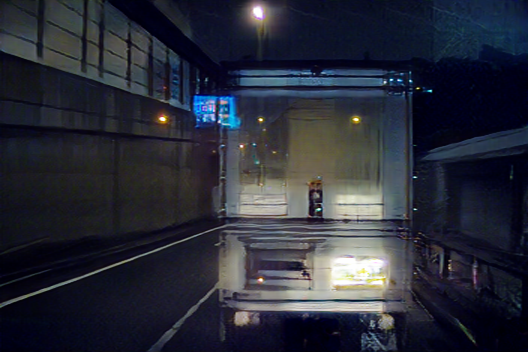}}
  \caption{\#blocks = 3}\end{subfigure}\hfill
  \begin{subfigure}{0.247\linewidth}
   {\includegraphics[width=1\linewidth]{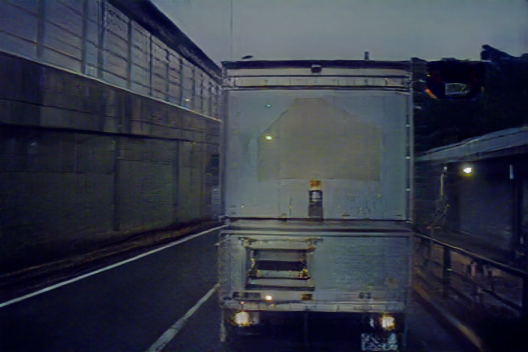}}
  \caption{\#blocks = 6}\end{subfigure}\hfill
  \begin{subfigure}{0.247\linewidth}
   {\includegraphics[width=1\linewidth]{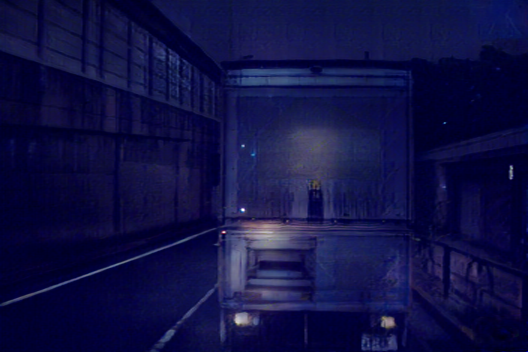}}
  \caption{\#blocks = 9}\end{subfigure}\hfill
\vspace{-5pt}
	\caption{\textbf{Visual results on variants of the number of Transformer blocks.} (a) input image and translated images (b,c,d) with different number of Transformer blocks for sunny$\rightarrow$night.}
	\vspace{-10pt}
	\label{fig:block_supp}
\end{figure*}

\paragraph{Effects of the Number of Transformer Blocks.}
We show qualitative and quantitative comparisons of the number of Transformers blocks. As described in~\tabref{tab:block} and~\figref{fig:block_supp}, the results using 6 and 9 blocks show almost close FID metric~\cite{heusel2017gans} and SSIM metric~\cite{wang2004image} scores and visual results, while the score of 3 blocks shows insufficient result. As smaller models have a lower parameter count, and a faster throughput, we choose 6 blocks for our architecture.\vspace{-10pt}

\begin{figure*}[h]
  \begin{subfigure}{0.247\linewidth}
  {\includegraphics[width=1\linewidth]{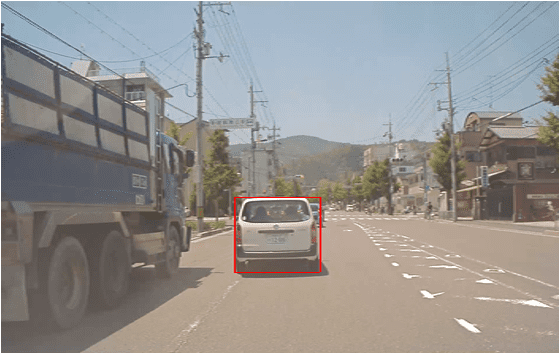}
  \includegraphics[width=1\linewidth]{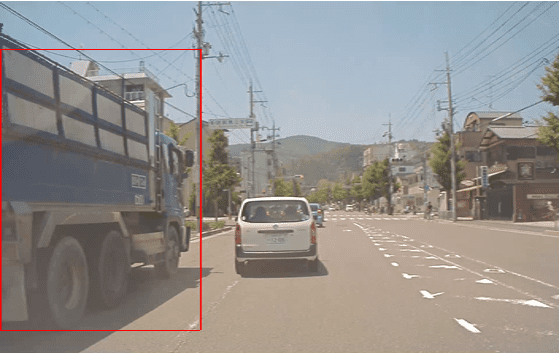}}
  \caption{Input image with instances }\end{subfigure}\hfill
  \begin{subfigure}{0.247\linewidth}
   {\includegraphics[width=1\linewidth]{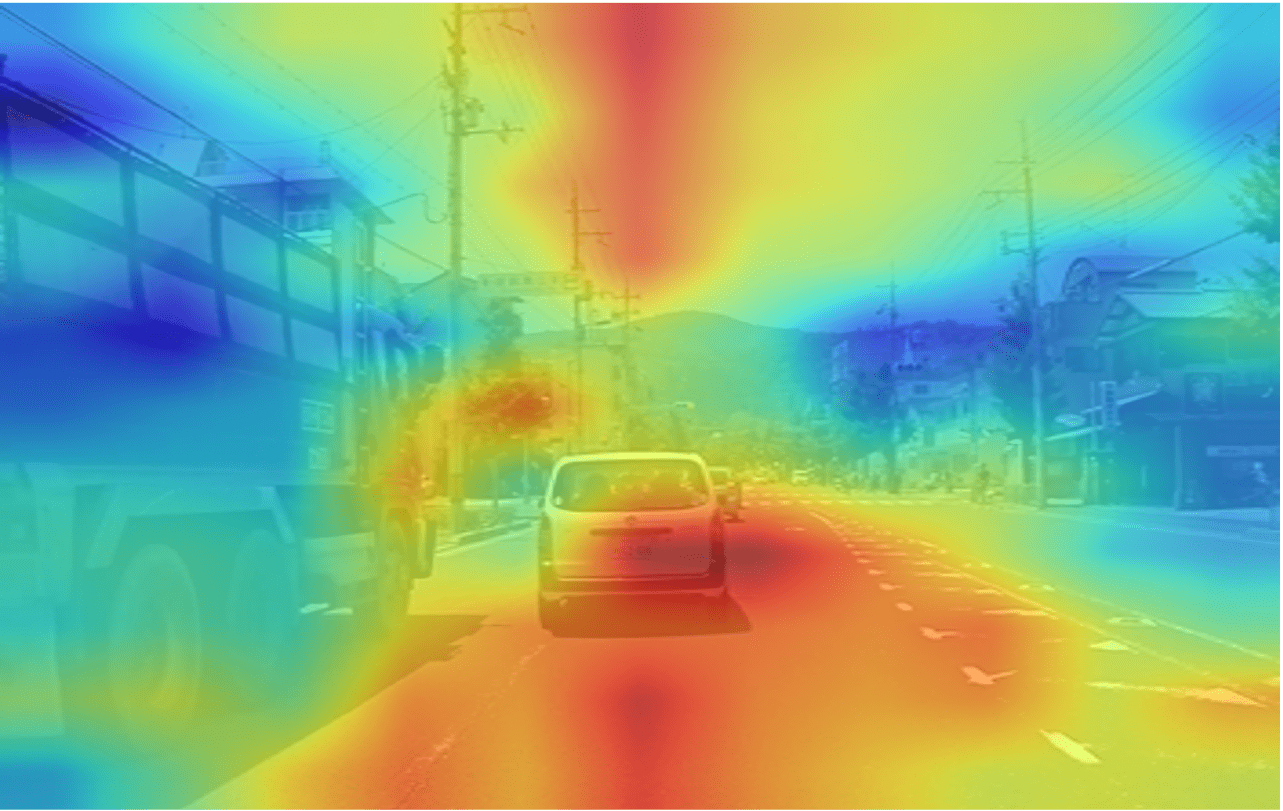}
   \includegraphics[width=1\linewidth]{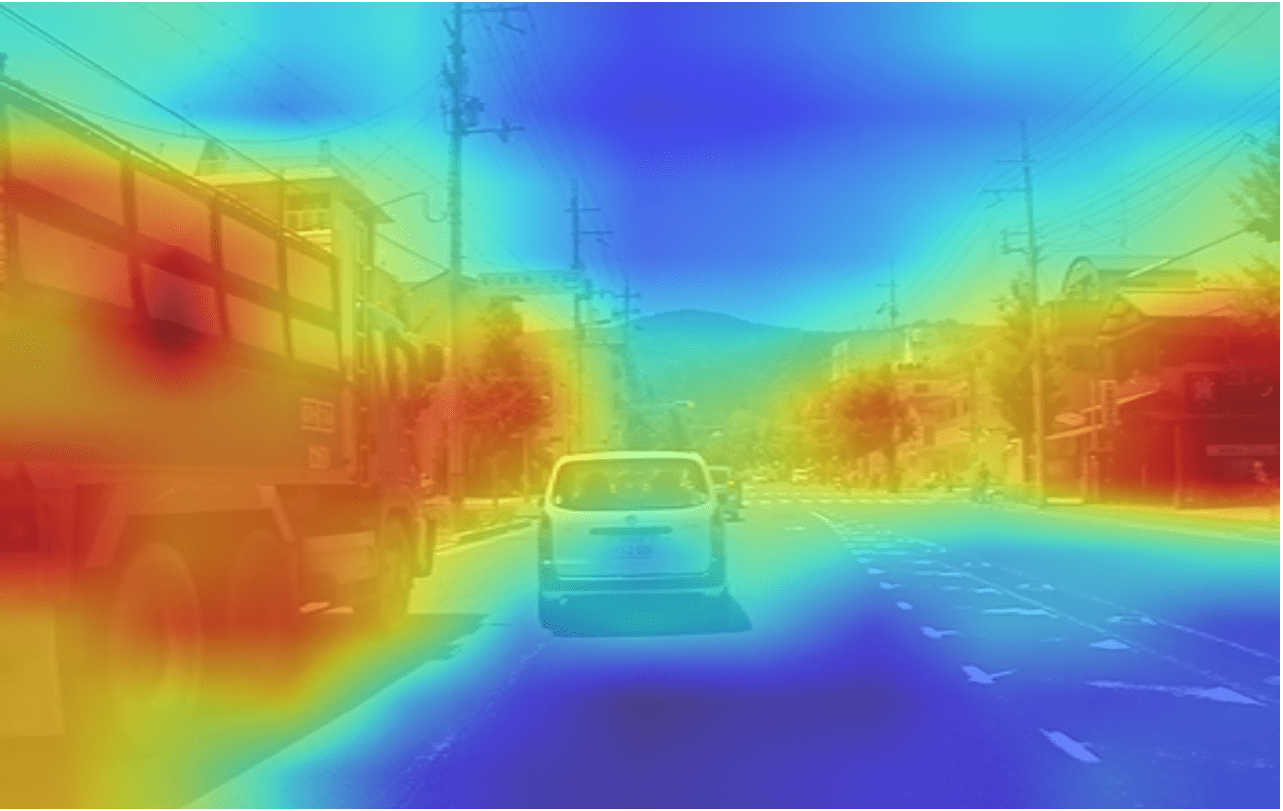}}
  \caption{\#heads = 1}\end{subfigure}\hfill
  \begin{subfigure}{0.247\linewidth}
   {\includegraphics[width=1\linewidth]{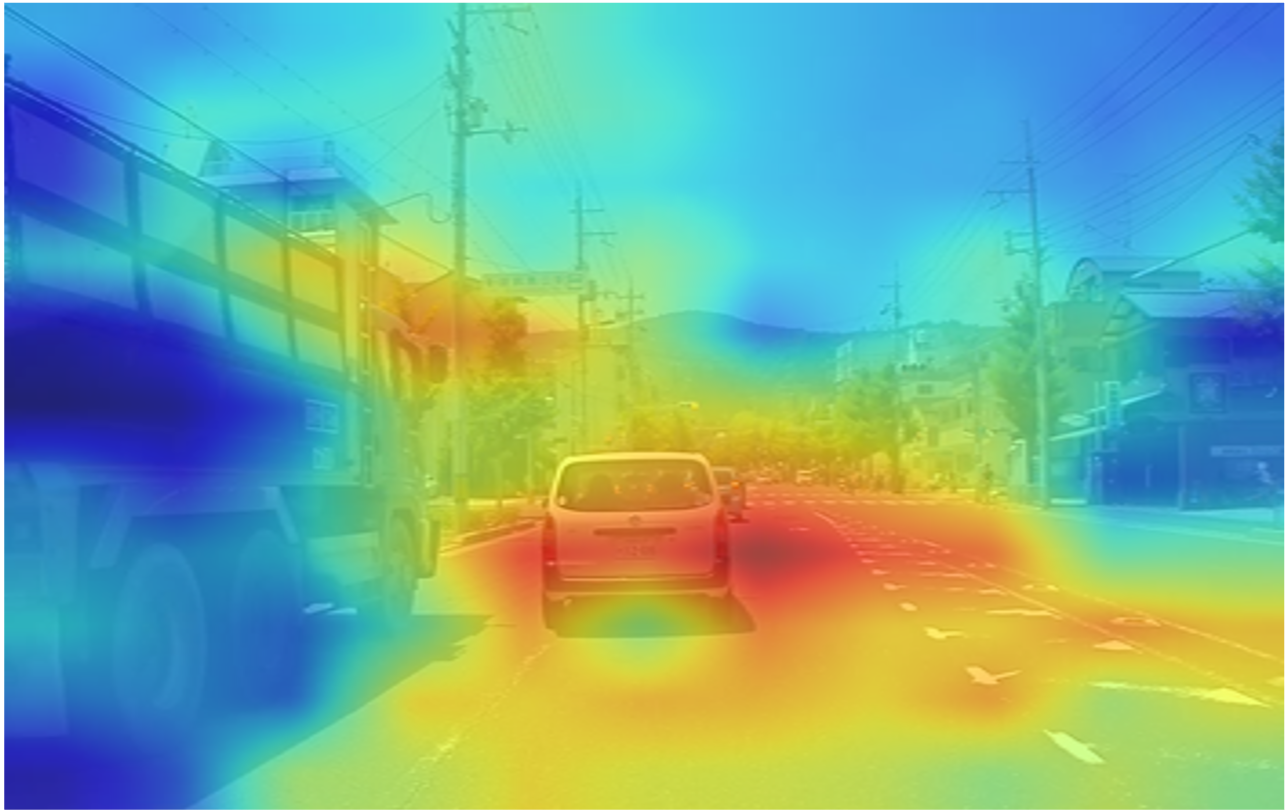}
   \includegraphics[width=1\linewidth]{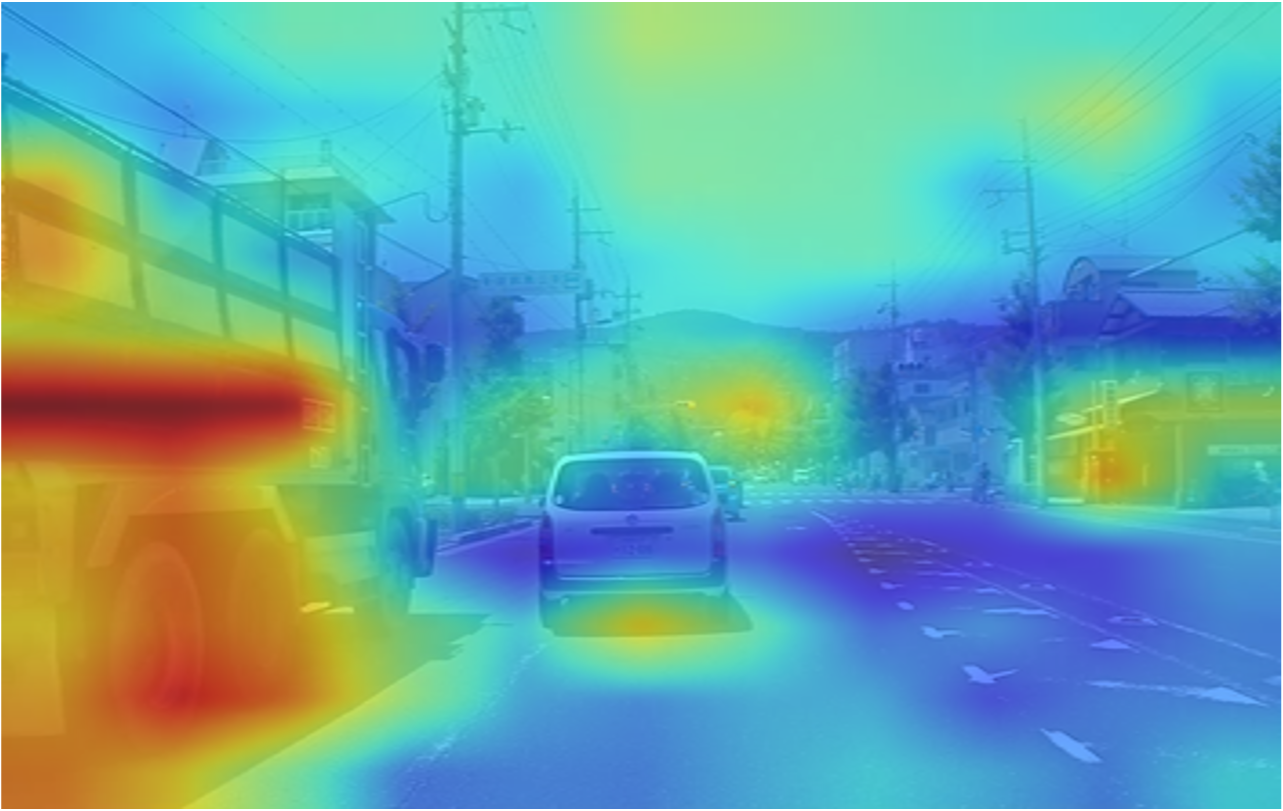}}
  \caption{\#heads = 4}\end{subfigure}\hfill
  \begin{subfigure}{0.247\linewidth}
   {\includegraphics[width=1\linewidth]{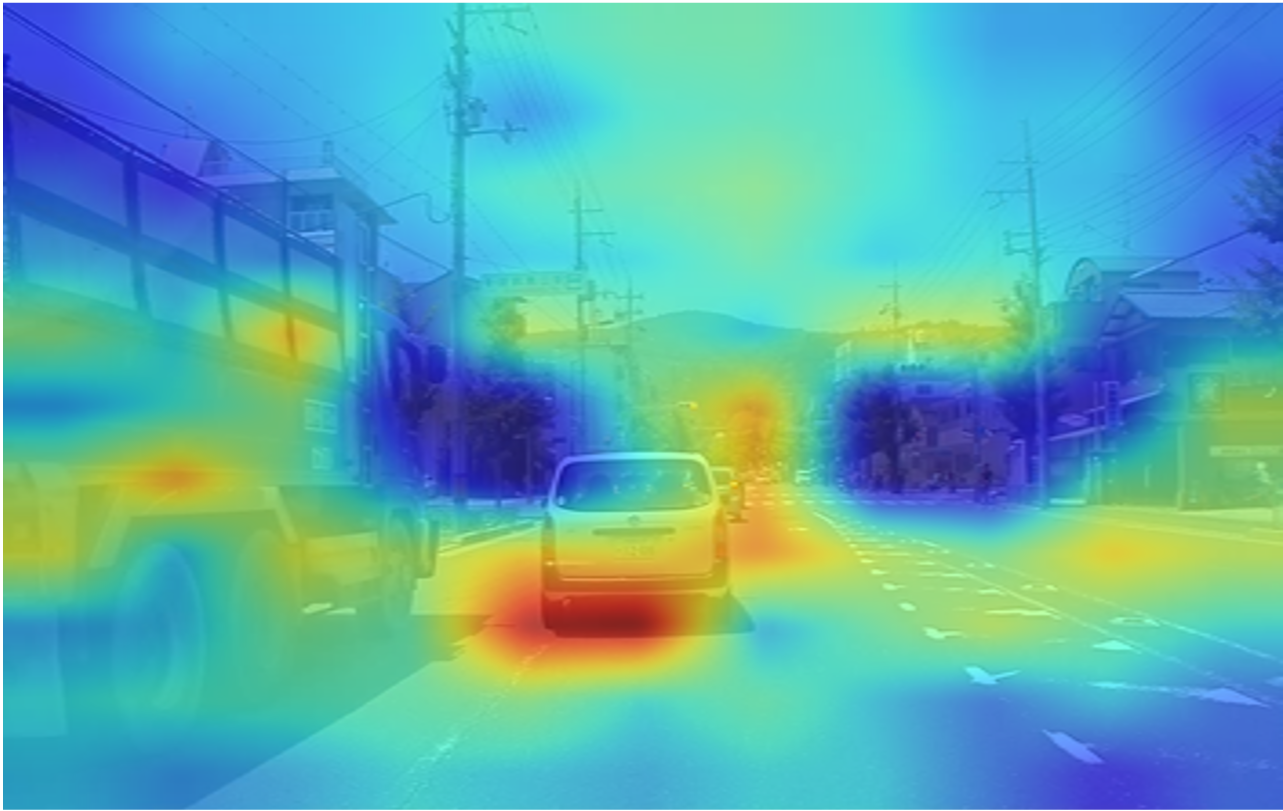}
   \includegraphics[width=1\linewidth]{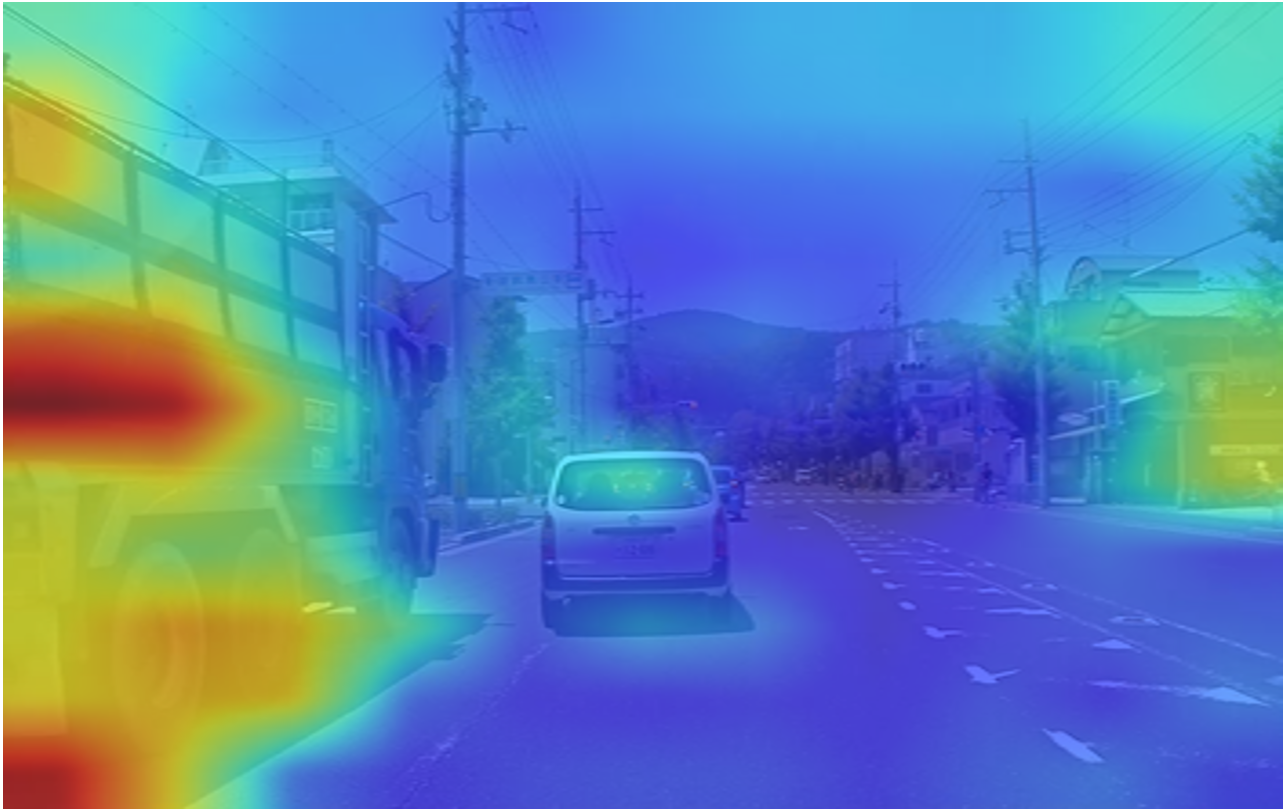}}
  \caption{\#heads = 8}\end{subfigure}\hfill
\vspace{-5pt}
	\caption{\textbf{Visualization of learned self-attention.} (from top to bottom) attention map for a car and a truc, respectively. (a) input image and following self-attention maps (b,c,d) for different number of heads.}
	\vspace{-10pt}
	\label{fig:att_supp}
\end{figure*}

\paragraph{Effects of the Number of Heads.} 
We analyze the effects of the number of heads in our model. We show the quantitative results in~\tabref{tab:block} and visualization of self-attention maps in~\figref{fig:att_supp} according to the number of heads. More heads tend to bring lower FID metric score and higher SSIM metric score with better self-attention learning, but the scores using 8 heads shows slight better results compared the scores using 4 heads. Thus, we decide to use 4 heads considering memory-efficiency.

\paragraph{Effects of Transformers.}
We additionally compare InstaFormer with CNN-based model in \tabref{tab:block}. While single head version of InstaFormer has a lower parameter count, it shows better performance in terms of FID metric compared to CNN-based model. This demonstrates that our outstanding performance is not due to its complexity.

\begin{table}[h]

\smallskip
\centering
\scalebox{0.8}
{
\begin{tabular}{l|cc|cccc}
\toprule
Variants & \#blocks  & \#heads & \#params &FID$\downarrow$& SSIM$\uparrow$ \\
\midrule
Less blocks&  3 & 4 &  37.776M  & 89.96 & 0.711 \\
Ours &  6 & 4 &  75.552M & 84.72 & 0.872 \\
More blocks& 9 & 4 &  113.329M & 85.28 & \textbf{0.879}\\\hline
Less heads&  6 & 1 &  4.732M & 89.17 & 0.738 \\
Ours &  6 & 4 &  75.552M & 84.72 & 0.872 \\
More heads&  6 & 8 &  302.100M & \textbf{81.92} & 0.873 \\
\midrule
CNN-based& 6 & - & 7.081M & 89.73 & 0.708 \\
\bottomrule
\end{tabular}}
\caption{\textbf{Effects of the number of blocks and heads in our InstaFormer architecture, providing quantitative evaluations with FID~\cite{heusel2017gans} and SSIM~\cite{wang2004image}.}
The only setting that vary across model is the number of Transformer blocks or heads, and we keep the others constant for sunny$\rightarrow$night on INIT dataset~\cite{shen2019towards}. Larger models tend to have a higher parameter count, and better FID~\cite{heusel2017gans} and SSIM~\cite{wang2004image} metric scores.
\label{tab:block}}
\end{table}

\end{document}